\pdfoutput=1
\documentclass{article}

     \usepackage[preprint, nonatbib]{neurips_2023}
\usepackage[numbers]{natbib}

\usepackage[utf8]{inputenc} % allow utf-8 input
\usepackage[T1]{fontenc}    % use 8-bit T1 fonts
\usepackage{hyperref}       % hyperlinks
\usepackage{url}            % simple URL typesetting
\usepackage{booktabs}       % professional-quality tables
\usepackage{amsfonts}       % blackboard math symbols
\usepackage{nicefrac}       % compact symbols for 1/2, etc.
\usepackage{microtype}      % microtypography
\usepackage{xcolor}         % colors

\usepackage{microtype}
\usepackage{graphicx}
\usepackage{svg}
\usepackage{caption}
\usepackage{subcaption}
\usepackage{pifont}% http://ctan.org/pkg/pifont
\usepackage{pbox}
\usepackage[ruled]{algorithm}
\usepackage{algorithmic}
\usepackage{fancyhdr,graphicx,amsmath,amssymb}
\usepackage{times}
\usepackage{wrapfig}
\usepackage{amsmath}
\usepackage{color}
\usepackage{tikz}
\usepackage{cancel}
\usepackage{overpic}
\usepackage{multirow}
\usepackage{amssymb}
\usepackage{mathtools}
\usepackage{amsthm}
\usepackage{nicefrac}

\theoremstyle{plain}
\newtheorem{theorem}{Theorem}[section]
\newtheorem{proposition}[theorem]{Proposition}

\theoremstyle{definition}

\theoremstyle{remark}

\usetikzlibrary{shapes,decorations,arrows,calc,arrows.meta,fit,positioning}
\tikzset{
    -Latex,auto,node distance =1 cm and 1 cm,semithick,
    state/.style ={ellipse, draw, minimum width = 0.7 cm},
    point/.style = {circle, draw, inner sep=0.04cm,fill,node contents={}},
    bidirected/.style={Latex-Latex,dashed},
    el/.style = {inner sep=2pt, align=left, sloped}}
\title{Leveraging sparse and shared feature activations for disentangled
representation learning}

\author{%
Marco Fumero \\
Sapienza, University of Rome 
\And
Florian Wenzel  \\
Amazon AWS 
\And
Luca Zancato  \\
Amazon AWS 
\And
Alessandro Achille  \\
Amazon AWS 
\And
 Emanuele Rodolà  \\
\hspace{-0.25cm} Sapienza, University of Rome 
\And
Stefano Soatto \\
Amazon AWS 
\And
Bernhard Schölkopf   \\
Amazon AWS 
\And
Francesco Locatello  \\
IST Austria }
\begin{document}

\maketitle

\begin{abstract}
Research on recovering the latent factors of variation of high dimensional data has so far focused on simple synthetic settings. Mostly building on unsupervised and weakly-supervised objectives, prior work missed out on the positive implications for representation learning on real world data. In this work, we propose to leverage knowledge extracted from a diversified set of supervised tasks to learn a common disentangled representation. Assuming that each supervised task only depends on an unknown subset of the factors of variation, we disentangle the feature space of a supervised multi-task model, with features activating sparsely across different tasks and information being shared as appropriate. Importantly, we never directly observe the factors of variations, but establish that access to multiple tasks is sufficient for identifiability under sufficiency and minimality assumptions.
We validate our approach on six
real world distribution shift benchmarks, and different data modalities (images, text), demonstrating how disentangled representations can be transferred to real settings.
\end{abstract}

\section{Introduction}
%\FL{I would turn the intro the other way around, we learn disentangled representations on image data under the assumption that we have multiple tasks relying on different factors of variations. This transfers well to new tasks. I see this has a very strong bias towards disentanglement that could be attacked. let's discuss} \MF{rephrased}
A fundamental question in deep learning is how to learn meaningful and reusable representation from high dimensional data observations \cite{bengio2013representation,lecun2015deep,Scholkopf2021Feb,schmidhuber1992learning}. A core area of research pursuing is centered on disentangled representation learning (DRL)~\cite{locatello2019challenging,bengio2013representation,higgins2016beta} where the aim is to learn a representation which recovers the factors of variations (FOVs) underlying the data distribution. Disentangled representations
are expected to contain all the information present in the data in a compact
and interpretable structure \cite{kulkarni2015deep,chen2016infogan} while being independent from a particular task \cite{goodfellow2009measuring}. It has been argued that separating information into interventionally independent factors~\cite{Scholkopf2021Feb} can enable robust downstream predictions, which was partially validated in synthetic settings~\cite{dittadi2020transfer,locatello2020weakly}. Unfortunately, these benefits did not materialize in real world representations learning problems, largely limited by a lack of scalability of existing approaches.

In this work we focus on leveraging knowledge from different task objectives to learn better representations of high dimensional data, and explore the link with disentanglement and out-of-distribution (OOD) generalization on real data distributions.  Representations learned from a large diversity of tasks are indeed expected to be richer and generalize better to new, possibly out-of-distribution, tasks. However, this is not always the case, as different tasks can compete with each other and lead to weaker models.  This phenomenon, known as negative transfer \cite{marx2005transfer,wang2019characterizing} in the context of transfer learning or task competition \cite{standley2020tasks} in multitask learning, happens when a limited capacity model is used to learn two different tasks that require expressing high feature variability and/or coverage.
Aiming to use the same features for different objectives makes them noisy and often increases the sensitivity to spurious correlations \cite{hu2022improving,geirhos2020shortcut,beery2018recognition}, as features can be both predictive and detrimental for different tasks. %For example, jointly solving a localization (e.g. classifying forest versus water environments) and a recognition task (e.g. classifying the type of bird present in the scene) require mutually detrimental features that can also be spuriously correlated with each other~\cite{sagawa2020investigation}.
%\looseness=-1Consider the example of jointly solving a localization (e.g. classifying forest versus water environments) and a recognition task (e.g. classifying the type of bird present in the scene)~\cite{sagawa2020investigation}. Localization requires marginalizing out the identity, whereas recognition requires marginalizing the location, making the features that are informative for one task detrimental to the other. Forcing the model to capture both increases the (already high~\cite{sagawa2020investigation}) chance of the model using spurious correlations to make predictions for either task.
Instead, we leverage a diverse set of tasks and assume that each task only depends on an unknown subset of the factors of variation. We show that disentangled representations naturally emerge without any annotation of the factors of variations under the following two representation constraints:
\begin{itemize}
%My take on the properties (MARCO):
% feature suffiency: The representation is sufficient in the sense that is composed at least of the features needed to solve each task in the task distribution, independently.
%fature minimality: The representation is \emph{minimal} in the sense that duplicated or splitted features are avoided
\item	\emph{Sparse sufficiency}: Features should activate sparsely with respect to tasks.
%. i.e. a single task should leverage only a sparse set of features to have a low prediction error. The representation is sufficient in the sense that is composed at least of the features needed to solve each task in the task distribution, independently.
The representation is \emph{sparsely sufficient} in the sense that any given task can be solved using few features. 
\item	\emph{Minimality}: 
Features are maximally shared across tasks whenever possible.
% i.e., a single feature should be reused if it is useful for solving a particular task.
% i.e. a single feature should activate across all tasks if it leads to higher overall accuracy. 
% if it leads to a higher overall accuracy. 
%There are no duplicated, unused, or split features.
The representation is \emph{minimal} in the sense that features are encouraged to be reused, i.e., duplicated or split features are avoided. % across all tasks.
\end{itemize}

These properties are intuitively desirable to obtain features that (i) are disentangled w.r.t. to the factors of variations underlying the task data distribution (which we also theoretically argue in Proposition~\ref{thm:ident}), (ii) generalize better in settings where test data undergo distribution shifts with respect to the training distributions, and (iii) suffer less from problems related to negative transfer phenomena.
To learn such representations in practice, we implement a meta learning approach, enforcing feature sufficiency and sharing with a \emph{sparsity} regularizer and an entropy based \emph{feature sharing} regularizer, respectively, incorporated in the base learner.
Experimentally, we show that our model learns meaningful disentangled representations that enable strong generalization on real world data sets.
%Experimentally, we show that our model is able to perform very well on OOD generalization benchmarks on real data, \FL{showing that disentangled representations enable stronger generalization on real world data distributions.}
Our contributions can be summarized as follows:
\begin{itemize}
\item We demonstrate that is possible to learn disentangled representations leveraging knowledge from a distribution of tasks. For this, we propose a meta learning approach to learn a feature space from a collection of tasks while incorporating our sparse sufficiency and minimality principles favoring task specific features to coexist with general features.

% incorporating inductive biases that favor task specific features to cohexist with general features.
%\item We show that our approach entails a theoretically grounded causal interpretation, showing    promising directions for causal representation learning. 
\item	Following previous literature, we test our approach on synthetic data, validating in an idealized controlled setting that our sufficiency and minimality principles lead to disentangled features w.r.t. the ground truth factors of variation, as expected from our identifiability result in Proposition~\ref{thm:ident}. 
%\item	We propose a novel metric for task distance, which is obtained a direct consequence of the inductive biases that we enforce on the feature space.
\item	We extend our empirical evaluation to non-synthetic data where factors of variations are not known,  and show that our approach generalizes well out-of-distribution on different domain generalization and distribution shift benchmarks. % \FL{While disentanglement cannot be measured explicitly due to lack of ground-truth, we still observe that different tasks leverage sparsely sufficient and minimal features, hinting to a meaningful feature disentanglement.}%In particular, we observe the importance of feature minimality when the test distribution is far from the training distribution.
\end{itemize}

\section{Method}
% Given a set of observational data $\mathbf{x} \in \mathcal{X}$ and a distribution of tasks $\mathcal{P}_t$ defined on $\mathcal{X}$,
Given a distribution of tasks $t \sim \mathcal{T}$ and data $(\mathbf{x_t}, y_t) \sim \mathcal{P}_t$ for each task $t$,
we aim to learn a disentangled representation $g(\mathbf{x}) = \hat{\mathbf{z}} \in \hat{\mathcal{Z}}\subseteq\mathbb{R}^M$, which generalizes well to unseen tasks.  
We learn this representation $g$ by imposing the sparse sufficiency and minimality inductive biases. %: (i) \emph{sparse sufficiency}, for each task $t$ only a sparse subset of the features in $\hat{\mathcal{Z}}$ is activated for each task $t$, and (ii) \emph{minimality}, features in $\hat{\mathcal{Z}}$ are maximally shared between the tasks $t \sim \mathcal{T}$ encouraging reusage of features. Based on these two principles, we propose the following method.

\subsection{Learning sparse and shared features}
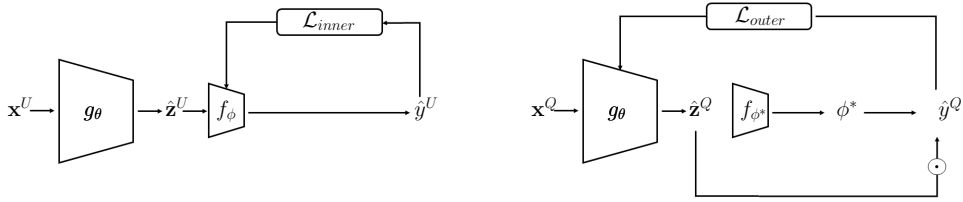
\begin{figure}[t]
\centering
\resizebox{\linewidth}{!}{
\begin{tabular}{ll}
\begin{overpic}[width=\linewidth,trim={4cm 4cm 4cm 4cm},clip]{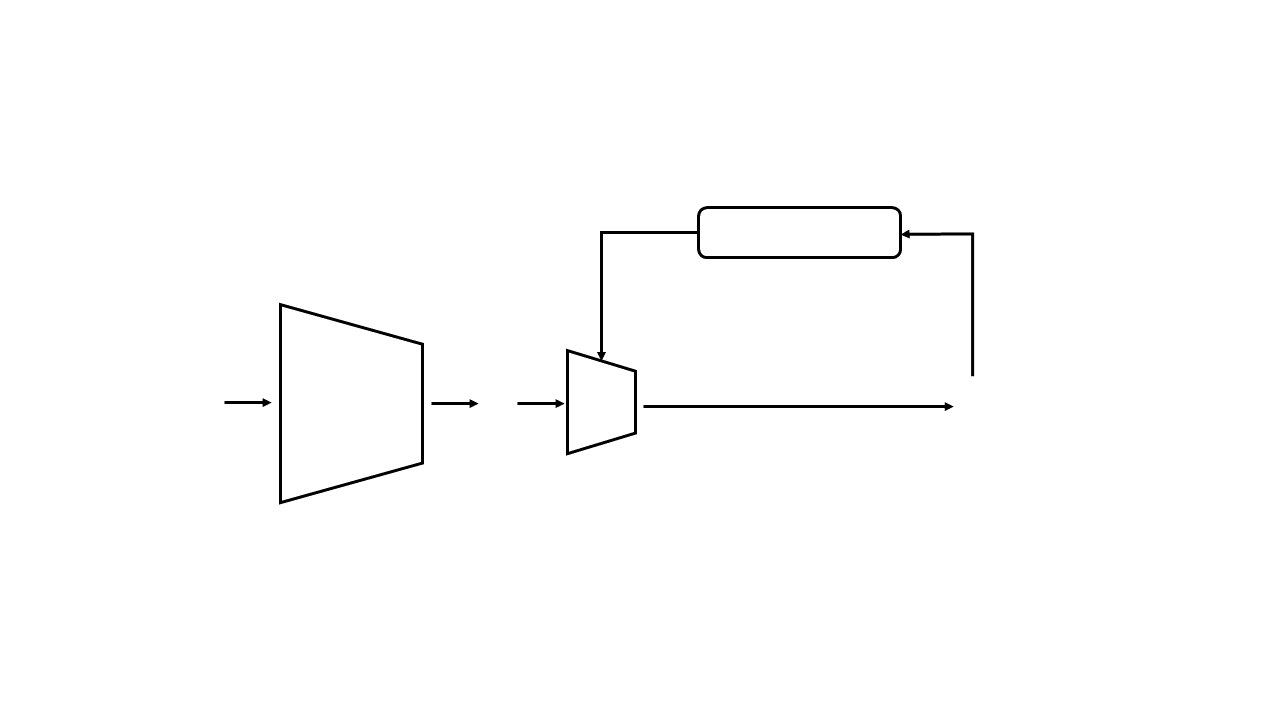}
\put(3,16){\LARGE$\mathbf{x}^U$}
\put(18,16){\LARGE$g_\theta$}
\put(44,16){\LARGE$f_\phi$}
\put(34,16){\LARGE$\hat{\mathbf{z}}^U$}
\put(18,16){\LARGE$g_\theta$}
\put(83,16){\LARGE$\hat{y}^U$}
\put(61,33.5){\LARGE$\mathcal{L}_{inner}$}
\end{overpic}&
\begin{overpic}[width=\linewidth,trim={4cm 4cm 4cm 4cm},clip]{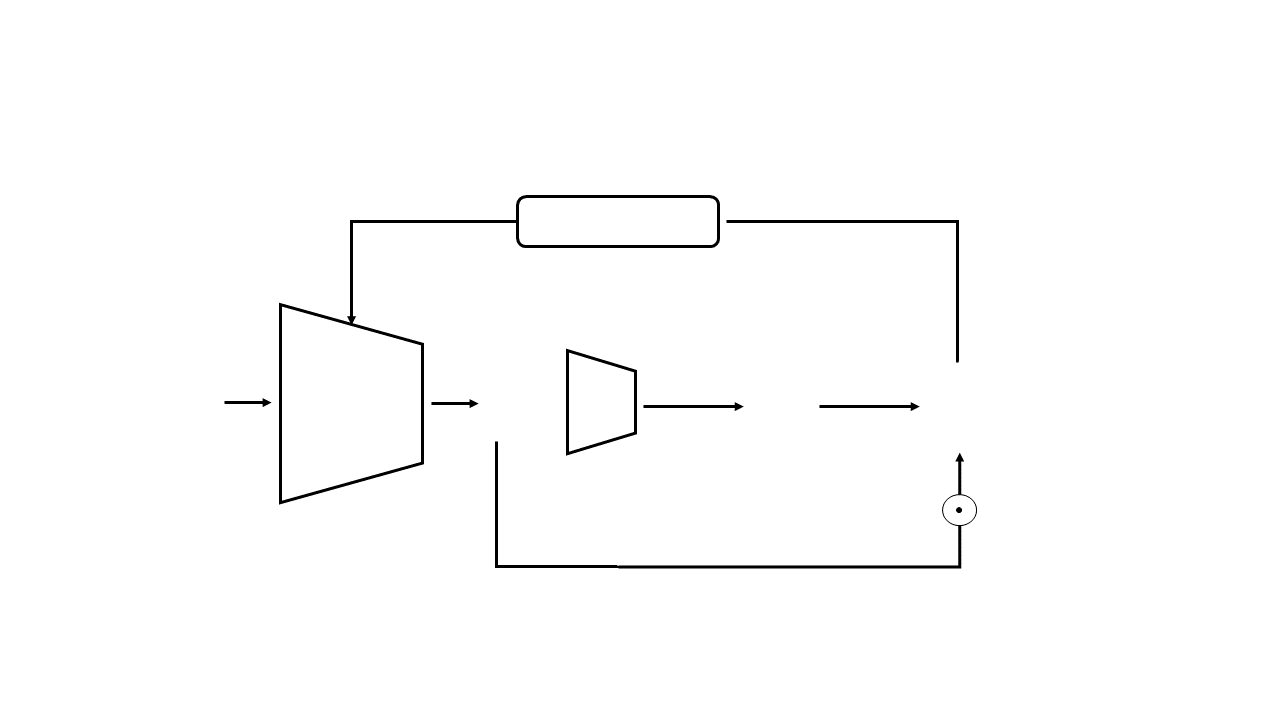}
\put(3,16){\LARGE$\mathbf{x}^Q$}
\put(18,16){\LARGE$g_\theta$}
\put(44,16){\LARGE$f_{\phi^*}$}
\put(63,16){\LARGE$\phi^*$}
\put(34,16){\LARGE$\hat{\mathbf{z}}^Q$}
\put(18,16){\LARGE$g_\theta$}
\put(83,16){\LARGE$\hat{y}^Q$}
\put(43,34.5){\LARGE$\mathcal{L}_{outer}$}
\end{overpic}
\end{tabular}}
\caption{\emph{Model scheme}: Illustrations of the (\textit{Top}) the inner loop stage and outer loop following the steps of the algorithmic procedure described in Section 
\ref{sec:app_algo} in the Appendix.} %\FL{Expand}}
\label{figure:arch}
\vspace{-0.5cm}
\end{figure}

Our architecture (see Figure~\ref{figure:arch}) is composed of a backbone module $g_\theta$ that is shared across all tasks and a separate linear classification head $f_{\phi_t}$, which is specific to each task $t$. The backbone is responsible to compute and learn a general feature representation for all classification tasks. The linear head solves a specific classification problem for the task-specific data $(\mathbf{x_t}, y_t) \sim \mathcal{P}_t$ in the feature space $\hat{\mathcal{Z}}$ while enforcing the feature sufficiency and minimality principles. %$ on $\hat{\mathbf{z}}_t$.
Adopting the typical meta-learning setting \cite{hospedales2020meta}, the backbone module $g_\theta$ can be viewed as the \textit{meta learner} while the task-specific classification heads $f_{\phi_t}$ can be viewed as the \textit{base learners}. In the meta-learning setting we assume to have access to samples for a new task given by a \textit{support set} $U$, with elements $(\mathbf{x}^U,y^U) \in U$. These samples are used to fit the linear head $f_{\phi^*}$ leading to the optimal feature weights for the given task. For a \textit{query} $\mathbf{x}^Q \in Q$, the prediction is obtained by computing the forward pass $\hat{y} = f_{\phi^*}(g_\theta(\mathbf{x}^Q))$.
 % We describe the details of our method in the following.

\paragraph{Enforcing feature minimality and sufficiency.}
%\FW{Introduce the regularizer first and explain why/how it relates to our principles.}\MF{Done}

To solve a task in the feature space $\hat{\mathcal{Z}}$ of the backbone module we impose the following regularizer $Reg(\phi)$
% along with the standard cross entropy loss term
on the classification heads $f_\phi$ with parameter $\phi \in  \mathbb{R}^{T \times M \times C}$, where $T$ is the number of tasks, $M$ the number of features, and $C$ the number of classes.
% The parameters $\phi \in \mathbb{R}^{T\times M \times C}$ for a set of $T$ tasks, with $M$ features and $C$ classes.
The regularizer is responsible for enforcing the feature minimality and sufficiency properties. It is composed of the weighted sum of a sparsity penalty $Reg_{L1}$ and an entropy-based feature sharing penalty: $Reg_{sharing}$
%, weighted respectively by scalar weights $\alpha$ and $\beta$:
\begin{equation}
    Reg(\phi) = \alpha Reg_{L_1}(\phi) +\beta Reg_{sharing}(\phi), \label{eq:regularizer}
\end{equation}
with scalar weights $\alpha$ and $\beta$.
The penalty terms are defined by:
\begin{align}
& Reg_{L_1} (\phi) = \frac{1}{TC} \sum_{t,c,m} | \phi_{t,m,c} | \\
& Reg_{sharing} (\phi) = H(\tilde{\phi}_m)=-\sum_m \tilde{\phi}_m log(\tilde{\phi}_m)
\end{align}
% where $\tilde{\phi}_m = \sum_m (\frac{1}{TC}\frac{ \sum_{t,c} |\phi_{t,c} |)}{ \sum_{t,c,m} |\phi_{t,c,m} |})$.
where $\tilde{\phi}_m = \frac{1}{TC}\frac{ \sum_{t,c} |\phi_{t,c,m} |}{ \sum_{t,c,m} |\phi_{t,c,m} |}$ are the normalized classifier parameters.
Sufficiency is enforced by a sparsity regularizer given by the $L_1$-norm, which constrains classification head to use only a sparse subset of the features. Minimality is enforced by the feature sharing term: minimizing the entropy of the distribution of feature importances (i.e. normalized $|\phi_{t}|$) averaged across a mini batch of $T$ tasks, leads to a more peaked distribution of activations across tasks. This forces features to cluster across tasks and therefore be reused by different tasks, when useful.We remark that different choices for the regularizers coming from the linear multitask learning literature (e.g. \cite{lozano2012multi,janati2019wasserstein,jajali:2010}) to enforce sparse sufficiency and minimality are indeed possibile.  We leave their exploration as a future direction.

\subsection{Training method}
\label{sec:training_method}
% \paragraph{Bi-level optimization problem.}
We train the model in meta-learning fashion by minimizing the test error over the expectation of the task distribution $t \sim \mathcal{T}$. This can be formalized as a \textit{bi-level optimization problem}. The optimal backbone model $g_{\theta^*}$ is given by the \textit{outer optimization problem}:
\begin{equation}
    \label{eq:outer_problem}
    \min_\theta \mathbb{E}_t [\mathcal{L}_{outer}(f_{\phi^*}(g_\theta (\mathbf{x}_t^Q),y_t^Q))],
\end{equation}
where $f_{\phi^*}$ are the optimal classifiers obtained from solving the \textit{inner optimization problem}, and ($\mathbf{x}_t^Q,y_t^Q) \in Q_t$ are the test (or query) datum from the query set $Q_t$ for task $t$.
%and corresponding label
Let $U_t$ be the support set with samples ($\mathbf{x}_t^U,y_t^U) \in U$ for task $t$, where typically the support set is distinct from the query set, i.e., $U \cap Q = \emptyset$. The optimal classifiers $f_{\phi^*}$ are given by the \textit{inner optimization problem}:
\begin{align}
    \label{eq:inner_problem}
    \min_\phi \frac{1}{T}\sum_t \mathcal{L}_{inner} (\hat{y}_t^U,y_t^U) + Reg(\phi),
    % & \textit{  where:   } \ \  \hat{y}_t^U= f_\phi(g_\theta(\mathbf{x}_t^U))
\end{align}
where $\hat{y}_t^U= f_\phi(g_\theta(\mathbf{x}_t^U)$. For both the inner loss $\mathcal{L}_{inner}$ and outer loss $\mathcal{L}_{outer}$ we use the cross entropy loss. %loss with respect to the support labels $y_t^U$ and $Reg(\phi)$ is defined \eqref{eq:regularizer}. 

\textbf{Task generation.} Our method can be applied in a standard supervised classification setting where we construct the tasks on the fly as follows. 
% we sample  minibatches of $T$ tasks (or episodes) from  the distribution $\mathcal{P}_t$.
% Each task $t$ is a classification task on a $C$-way problem with $|S|$ samples and labels, referred as \emph{support set}.
% To sample a task in practice, 
We define a task $t$ as a $C$-way classification problem. We first select a random subset of $C$ classes from a training domain $D_{train}$ which contains $K_{train}$ classes. For each class we consider the corresponding data points and select a random support set $U_t$ with elements $(\mathbf{x}_t^U,y^U) \in U$ and a disjoint random query set $Q_t$ with elements $(\mathbf{x}_t^Q,y^Q) \in Q_t$. 
% Therefore $t:=(S_t,Q_t)$ and $S_t\cap Q_t=\emptyset$.
% To denote samples and labels coming from the support (query) set of a given task we use the following notation: $\mathbf{x}_t^U,y_t^U={(x,y)\in S_t}$,  (equivalently, $\mathbf{x}_t^Q,y_t^Q={(x,y) \in Q_t}$). 
% The same procedure can be repeated for the validation and test domains $D_{val}$ and $D_{test}$.

% The support samples  $\mathbf{x}_t^U,y_t^U \in S_t$ are used to update the linear head  coefficients $\phi$ in the inner loop stage of the optimization problem, while the query samples  $\mathbf{x}_t^Q,y_t^Q={(x,y) \in Q_t}$ are be used to update the backbone weights $\theta$ in the outer loop.

\textbf{Algorithm.}
\looseness=-1In practice we solve the bi-level optimization problem~\eqref{eq:outer_problem} and \eqref{eq:inner_problem} as follows. In each iteration we sample a batch of $T$ tasks with the associated support and query set as described above. First, we use the samples from the support set $S_t$ to fit the linear heads $f_\phi$ by solving the inner optimization problem~\eqref{eq:inner_problem} using stochastic gradient descent for a fixed number of steps. Second, we use the samples from the query set $Q_t$ to update the backbone $g_\theta$ by solving the outer optimization problem~\eqref{eq:outer_problem} using implicit differentiation~\cite{blondel2021efficient,griewank2008evaluating}. Since the optimal solution of the linear heads $\phi^*$ depend on the backbone $g_\theta$, a straightforward differentiation w.r.t. $\theta$ is not possible. We remedy this issue by using the approximation strategy of \cite{geng2022training} to compute the implicit gradients. The algorithm is summarized in section \ref{sec:app_algo} of the Appendix.%\FW{Marco, please add}.

\subsection{Theoretical analysis}\label{sec:theory}
% \FW{This needs an introduction/motivation. Why are we doing the theoretical analysis? What is the result/implication of it?}\MF{Done}

\begin{figure}[t]%{R}{0.5\linewidth}
     \centering
     \includegraphics[width=0.35\linewidth,trim={0 0 0 0cm},clip]{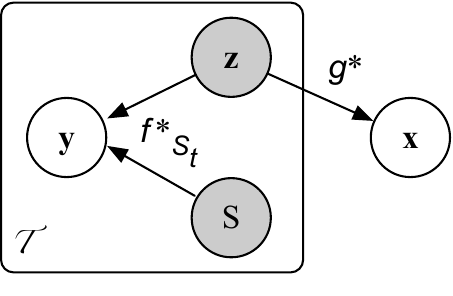}
%       \begin{tikzpicture}

%     \node[state,fill=gray] (1) {$S$};
   
%     \node[state,fill=gray] (3) [above =of 1] {$\mathbf{z}$};
%     %\node[state] (4) [above right =of 1,xshift=-0.3cm,yshift=-0.3cm] {$X_\bot$};
%     \node[state] (4) [ right =of 3,yshift=-1cm] {$\mathbf{x}$};
%     \node[state] (5) [left =of 1,yshift=0.5cm] {$\mathbf{y}$};

% \node (r1) [draw=none, thick,fill=none, minimum width=1.2cm,minimum height=3.5cm]{};
%     \path (1) edge node[above] {} (5);
%     \path (3) edge node[above] {} (5);
%     %\path[bidirected] (2) edge[bend left=60] node[above] {$\epsilon_{xy}$} (3);
%     \path (3) edge node[el,above] {$g^*$} (4);
%     %\path (1) edge node[el,above] {} (4);
% \end{tikzpicture}
     \caption{Assumed causal generative model: the gray variables are unobserved. Observations $\mathbf{x}$ are generated by some unknown mixing of a set of factors of variations $\mathbf{z}$. Additionally, we observe a distribution of supervised tasks, only depending on a subset of factors of variations indexed by $S$.}%\FL{you can use wrapfigure here to save space} \MF{solved}}
     \label{fig:causal_model}
 \end{figure}

We analyze the implications of the proposed minimality and sparse sufficiency principles and show in a controlled setting that they indeed lead to identifiability. 
As outlined in Figure~\ref{fig:causal_model}, we assume that there exists a set of independent latent factors $\mathbf{z} \sim \prod_{i=1}^d p(z_i)$ that generate the observations via an unknown mixing function $\mathbf{x} = g^*(\mathbf{z})$. Additionally, we
assume that the labels $y_t$ for a task $t$ only depend on a subset of the factors
indexed by $S_t \sim P(S)$, where $S$ is an index set on $\mathbf{z} \in \mathcal{Z}$, via some unknown mixing function
$y_t = f_{t}^* (\mathbf{z})$ (potentially different for different tasks).
% given specific assumptions on the task distribution $\mathcal{T}$.
%With the following result we are able to demonstrate that disentanglement arises naturally from the sufficiency and minimality principles we assumed.
% Specifically, given the distribution $\mathcal{P}_t$ of $t$ supervised tasks, where $(\mathbf{x}_t,y_t) \sim \mathcal{P}_t $ we aim to learn a representation $\hat{\mathbf{z}}_{t}=g(\mathbf{x}_{t})$ which is disentangled w.r.t. ground truth factors of variations $\mathbf{z}$.%should generalize well to unseen tasks. 
 %We assume the existence of the following latent generative model on $\mathbf{z}$ (depicted in Figure \ref{fig:causal_model}):
%\begin{align}&p(\mathbf{z})=\prod_ip(z_i) \ \  \ \ \ \ \ \ \ \ \ \ \   &&S \sim p(S) \\ &\mathbf{x} = g^*(\mathbf{z}) && y=f^*(\mathbf{z} |S)
%\end{align}
%where $S$ is an index set indicating the support on $Z$  for each task $t \sim \mathcal{P}_t$, i.e. which the factors of variation are used by each task and $f^*$ is the task generating function.
%Our goal is to show that is possible to recover a representation $\hat{\mathbf{z}}$ axis aligned, component wise transformation of $\mathbf{z}$, from observations of $p(X,Y)$ alone. Importantly we don't assume  the dimensionality of $\hat{\mathbf{z}}$ to match the one of the true $\mathbf{z}$, meaning that we can recover the factors of variation from an overestimate of $|Z|$.
We formalize the two principles that are imposed on $f^*$ by:
\begin{enumerate}
    \item \textit{sufficiency}: $f_t^*=f_t^*|_{S_t} \ \ \text{for} \  S_t \sim p(\mathcal{S})$
    \item \textit{minimality}: $\not \exists S' \neq S_t \subset  \mathcal{S} \text{  s.t.  } f_t^*|_{S'}=f_t^* $,
\end{enumerate}
where $f|_{S_t}$ denotes that the input to a function $f$ is restricted to the index set given by $S_t$ (all remaining entries are set to zero). (1) states that $f_t^*$ only uses a subset of features, and (2) states that there are not be duplicate features.

\begin{proposition}\label{thm:ident}
Assume that
$g^*$ is a diffeomorphism (smooth with smooth inverse), $f^*$ satisfies the sufficiency and minimality properties stated above, and $p(S)$ satisfies: $p(S \cap S' = \{i\})> 0$ or  $p( \{i\} \in (S \cup S') -(S'\cap S))>0$. Observing unlimited data from $p(X,Y)$, it is possible to recover a representation $\hat{\mathbf{z}}$ that is an axis aligned, component wise transformation of $\mathbf{z}$. %In other words the distribution over learned features $q(\hat{\mathbf{z}})$ is disentangled in the sense that it matches the joint distribution $p(\mathbf{x},\mathbf{z},S,y) = p(\mathbf{x} |\mathbf{z}) \prod_ip(z_i)p(y|\mathbf{z},S)$, after marginalization on $S$ and $\mathbf{z}$, i.e. $q(\hat{\mathbf{z}})= \int q(\hat{\mathbf{z}}|\mathbf{x}) p(\mathbf{x}) d\mathbf{x}= \int\int q(\hat{\mathbf{z}}|\mathbf{x}) p(\mathbf{x} |\mathbf{z}) p(\mathbf{z}) d\mathbf{z} d\mathbf{x}$.

\end{proposition}
%\FL{need a discussion, I would maybe even call it corollary from the work of Seb. otherwise we keep it a thm, but we want to be humble here}\MF{agreed, what about a small discussion paragraph below?}
\textbf{Remarks:} Overall, we see this proposition as validation that in an idealized setting our inductive biases are sufficient to recover the factors of variation. 
%Out theoretical result shows that disentanglement is identifiable under sparse sufficiency and minimality assumptions. 
Note that the proof is non-constructive and does not entail a specific method. In practice, we rely on the same constraints as inductive biases that lead to this theoretical identifiability and experimentally show that disentangled representations emerge in controlled synthetic settings. %so it would be incorrect 
%In section \ref{sec:experiments}, we show that sufficient and minimal representations are useful for generalization on downstream tasks defined on real world data distribution. 
On real data, (1) we cannot directly measure disentanglement, (2) a notion of global ground-truth factors may even be ill-posed, and (3) the assumptions of Proposition~\ref{thm:ident} are likely violated. Still, sparse sufficiency and minimality yield some meaningful factorization of the representation for the considered tasks. %However, we will show that our representations satisfy our constraints and learn some factorization that is meaningful for the considered tasks. 

\looseness=-1\textbf{Relation to \cite{lachapelle2022synergies} and \cite{locatello2020weakly}}: Our theoretical result can be reconnected with concurrent work \cite{lachapelle2022synergies} and can be seen as a corollary with a different proof technique and slightly relaxed assumptions. 
The main difference is that our feature minimality allows us to also cover the case where the number of factors of variations is unknown, which we found critical in real world data sets (the main focus of our paper). Instead, they only assume sparse sufficiency, which is enough for identifiability if the ground-truth number of factors is known, but is not enough to recover high disentaglement when this is not the case (see Figure \ref{fig:betavsDCI}) and does not translate well to real data, see Table~\ref{tab:meta_learning} with the empirical comparison in Appendix \ref{sec:app_meta_learning}. Interestingly, their analysis also hints at the fact that our approach also benefits in terms of sample complexity on transfer learning downstream tasks. 
Our proof technique follows the general construction developed for multi-view data in~\cite{locatello2020weakly}, adapted to our different setting. Instead of observing multiple views with shared factors of variation, we observe a single task that only depend on a subset of the factors.

% with the feature sharing property playing a fundamental role. Our statement and proof we follow the construction of \cite{locatello2020weakly}, adapted it to our setting. While theoretically our result could be seen as a corollary of~\cite{lachapelle2022synergies} with a different proof technique, our paper is largely focused on scaling disentanglement to real world data, for which the assumption we relax is critical.

% \textbf{Transferring disentanglement to real data distributions}
% %\FW{Add a small paragraph on why we expect that disentangled respresentations help with multi-class learning and OOD generalization.}
% Out theoretical results how disentanglement emerges as natural consequence of sufficiency and minimality properties of the representation. We will show, in section \ref{sec:experiments}, that sufficient and minimal representations are useful for generalization on downstream tasks defined on real world data distribution. Despite we cannot directly measure disentanglement on real world data, as we don't have access to ground truth factors of variations, we can still validate how general are the representation that we learn, demonstrating their robustness to distribution shifts and spurious correlation in the data. %To this end, in experimental section \ref{sec:experiments} we show how representations that satisfies these properties and are learned from a distribution of tasks on real data, are indeed more robust to distribution shifts and spurious correlation in the data. Our focus in the paper is to bridge the gap between synthetic and real world, showing how disentangled representations induced by sufficiency and minimality properties transfer well to a real data setting,  achieving good OOD generalization. 

\section{Related work} 
\textbf{Learning from multiple tasks and domains.} Our method addresses the problem of learning a general representation across multiple and possibly unseen tasks~\cite{caruana1997multitask,zhang2018overview} and environments~\cite{zhou2021domain,gulrajani2020search,koh2021wilds,wortsman2022model,miller2021accuracy,wiles2021fine,muandet2013domain} that may be competing with each other during training~\cite{marx2005transfer,wang2019characterizing,standley2020tasks}. 
Prior research tackled task competition by introducing task specific modules that do not interact during training \cite{ParKilRojSch18,florence_21, flava_21}. While successfully learning specialized modules, these approaches can not leverage synergistic information between tasks, when present. On the other hand, our approach is closer to multi-task methods that aim at learning a generalist model, leveraging multi-task interactions \cite{uni_moe_22, ofasys22}.
Other approaches that leverage a meta-learning objective for multi-task learning have been formulated \cite{dhillon2019baseline,snell2017prototypical,lee2019meta,bertinetto2018meta}.
In particular, \cite{lee2019meta} proposes to learn a generalist model in a few-shot learning setting without explicitly favoring feature sharing, nor sparsity. Instead, we rephrase the multi-task objective function encoding both feature sharing and sparsity to avoid task competition.

Similar to prior work in domain generalization, we assume the existence of stable features for a given task~\cite{muandet2013domain,arjovsky2019,veicht2021,jiang2022,wang2022unified} 
and amortize the learning over the multiple environments. 
Differently than prior work, we do not aim to learn an invariant representation a priori. Instead, we learn sufficient and minimal features for each task, which are selected at test time fitting the linear head on them. In light of~\cite{gulrajani2020search}, one can interpret our approach as learning the final classifier using empirical risk minimization but over features learned with information from the multiple domains.

\textbf{Disentangled representations.}
Disentanglement representation learning \cite{bengio2013representation,higgins2016beta}
aims at recovering the factors of variations underlying a given data distribution. \cite{locatello2019challenging} proved that without any form of supervision (whether direct or indirect) on the Factors of Variation (FOV) is not possible to recover them. Much work has then focused on identifiable settings~\cite{locatello2020weakly,fumero2021learning} from non-i.i.d. data, even allowing for latent causal relations between the factors. Different approaches can be largely grouped in two categories. First, data may be non-independently sampled, for example assuming sparse interventions or a sparse latent dynamics~\cite{RIMs,lippe2022citris,brehmer2022weakly,yao2021learning,ahuja2020invariant,seigal2022linear,lachapelle2022disentanglement}.
Second, data may be non-identically distributed, for example being clustered in annotated groups \cite{hyvarinen2019nonlinear,khemakhem2020variational,sorrenson2020disentanglement,willetts2021don,lu2022invariant}.
Our method follows the latter, but we do not make assumptions on the factor distribution across tasks (only their relevance in terms of sufficiency and minimality). This is also reflected in our method, as we train for supervised classification as opposed to contrastive or unsupervised learning as common in the disentanglement literature.  The only exception is the work of~\cite{lachapelle2022synergies} discussed in Section~\ref{sec:theory}.

\section{Experiments}\label{sec:experiments}

We start by highlighting here the experimental setup of this paper along with its motivation.

\looseness=-1\textbf{Synthetic experiments.} We first evaluate our method on benchmarks from the disentanglement literature \cite{dsprites17,3dshapes18,reed2015deep,lecun2004learning} where we have access to ground-truth annotations and we can assess quantitatively how well we can learn disentangled representations. 
We further investigate how minimality and feature sharing are correlated with disentanglement measures (Section~\ref{sec:corr_dis_beta}) and how well our representations, which are learned from a limited set of tasks, generalize their composition. The purpose of these experiments is to validate our theoretical statement, showing that if the assumptions of Proposition~\ref{thm:ident} hold, our methods quantitatively recover the factors of variation. 

\textbf{Domain generalization.} On real data sets, we can neither quantitatively measure disentanglement nor are we guaranteed identifiability (as assumptions may be violated). Ultimately, the goal of disentangled representations is to learn features that lend themselves to be easily and robustly transferred to downstream tasks. Therefore, we first evaluate the usefulness of our representations with respect to downstream tasks subject to distribution shifts, where isolating spurious features was found to improve generalization in synthetic settings~\cite{dittadi2020transfer,locatello2020weakly}
To assess how robust our representations are to distribution shifts, we evaluate our method on domain generalization and domain shift tasks on six different benchmarks (Section~\ref{sec:dg}). In a domain generalization setting, we do not have access to samples coming from the testing domain, which is considered to be OOD  w.r.t. to the training domains. However, in order to solve a new task, our method relies on a set labeled data at test time to fit the linear head on top of the feature space. Our strategy is to sample data points from the training distribution, balanced by class, assuming that the label set $Y$ does not change in the testing domain, although its distribution may undergo subpopulation shifts.
%This sampling strategy is in line with what is highlighted in \cite{kirichenko2022last}, where it is shown that retraining the linear head of a deep classifier on a small set of balanced samples (w.r.t to minority groups in the training data) is sufficient to achieve robustness to spurious correlations in the test data. The main difference is that we typically don't assume to have labels on the minority groups in the training set and we just balance the sampling by the class label.

\textbf{Few-shot transfer learning.} 
Lastly, we test the adaptability of the feature space to new domains with limited labeled samples. For transfer learning tasks, we fit a linear head using the available limited supervised data. The sparsity penalty $\alpha$ is set to the value used in training; the feature sharing parameter $\beta$ is defaulted to zero unless specified.

\textbf{Experimental setting.}
%To have a fair comparison with other methods in the literature, we  simply substitute the backbone $f$ with the backbone $f'$ of the compared method (e.g. for ERM models, we detach the classification head, to separate the models into two modules) and then fit the linear head on the same data for our method and comparisons. 
To have a fair comparison with other methods in the literature, we adopt the standard experimental setting of prior work \cite{gulrajani2020search,koh2021wilds}. Hyperparameters $\alpha$ and $\beta$ are tuned performing model selection on validation set, unless specified otherwise. For comparison with baselines, we substitute our backbone with that of the baseline (e.g. for ERM models, we detach the classification head) and then fit a new linear head on the same data.
The linear head module trained at test time on top of the features is the same both for our and compared methods. 
Despite its simplicity, we report the ERM baseline for comparison in our experiments in the main paper, since it has been shown to perform best in average on domain generalization benchmarks \cite{gulrajani2020search,koh2021wilds}. We further compare with other consolidated approaches in the literature such as IRM \cite{arjovsky2019}, CORAL \cite{sun2016deep} and GroupDRO \cite{Sagawa2019} and include a large and comprehensive comparison with \cite{yan2020improve,blanchard2021domain,li2018learning,li2018domain,ganin2016domain,li2018deep,nam2021reducing,zhang2021adaptive,huang2020self,krueger2021out} in Appendix\ref{sec:app_full_Dg_exp}. Experimental details are fully described in Appendix~\ref{sec:exp_details}.

\subsection{Synthetic experiments} \label{sec:fovs_exp}

We start by demonstrating that our approach is able to recover the factors of variation underlying a synthetic data distribution like \cite{dsprites17}. For these experiments, we assume to have partial information on a subset of factors of variation $Z$, and we aim to learn a representation $\hat{\mathbf{z}}$ that aligns with them while ignoring any spurious factors that may be present. We sample random tasks from a distribution $\mathcal{T}$ (see Appendix \ref{sec:app_task_gen} for details) %$ We start by selecting a subset of the factors of variations $Z$ on which the task are defined, i.e. $supp(\mathcal{P}_S) \subset Z$. Then we sample tasks from $p(S)$, where each task $f^*: Z \rightarrow Y$ 
5and focus on binary tasks, with $Y=\{0,1\}$. For the \texttt{DSprites} dataset an example of valid task is \textit{``There is a big object on the left of the image''}. In this case, the partially observed factors (quantized to only two values) are the \textit{x position} and \textit{size}.
In Table~\ref{tab:disentanglement}, we show how the feature sufficiency and minimality properties enable disentanglement in the learned representations. We train two identical models on a random distribution of sparse tasks defined on FOVs, showing that, for different datasets
\cite{dsprites17,3dshapes18,lecun2004learning,reed2015deep}, the same model without regularizers achieves a similar in-distribution (ID) accuracy, but a much lower disentanglement. 

\begin{wrapfigure}{R}{0.5\linewidth}
    \vspace{-0.5cm}
    \begin{overpic}
    [width=\linewidth,trim={0.45cm 0 1.1cm 1.1cm},clip]{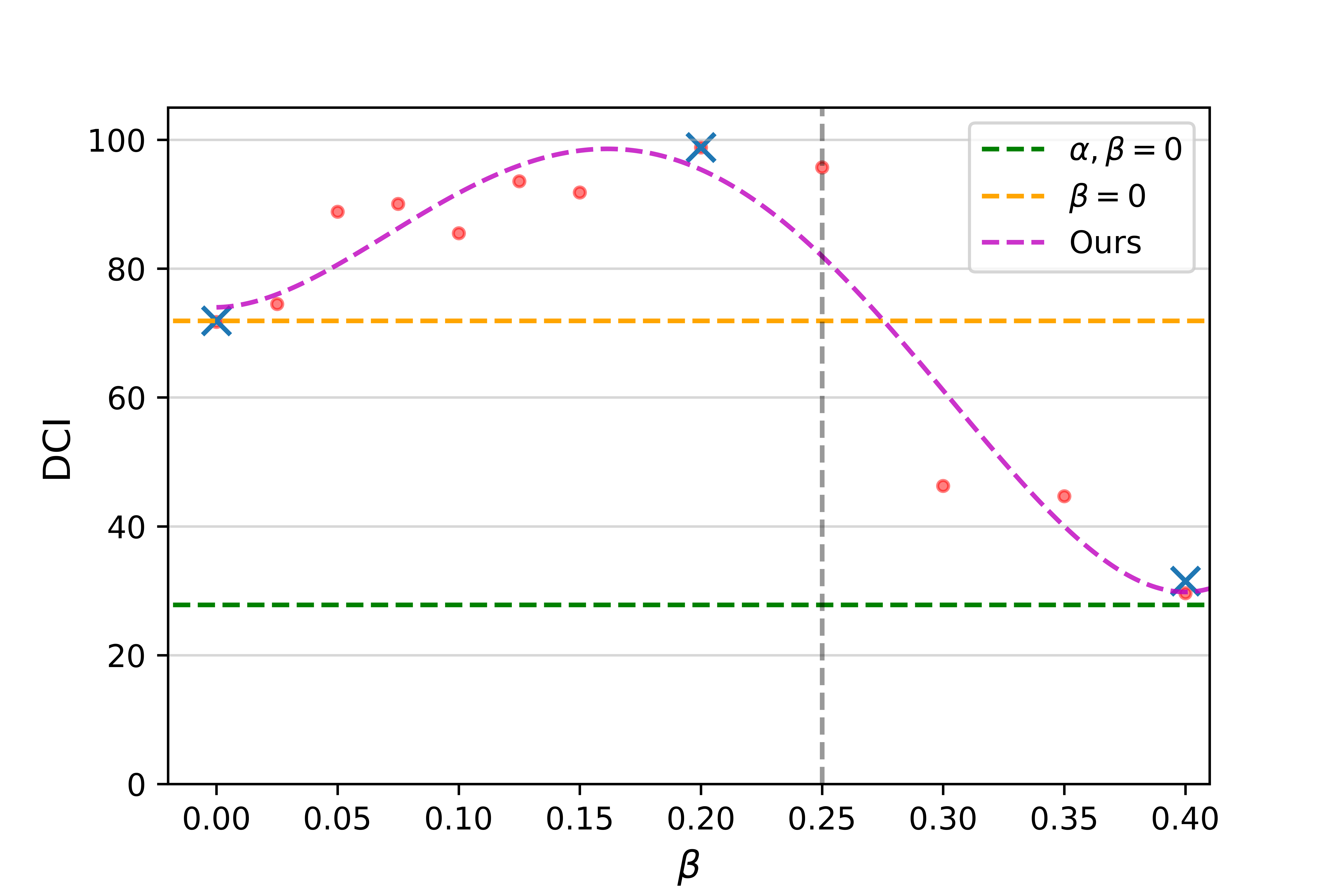}
    \put(91.3,57.2){\tiny\cite{lachapelle2022synergies}}
    %\put(20,20){\color{green}{-} \color{black} \footnotesize $\alpha=0,\beta=0$}
    \end{overpic}
    \vspace{-0.7cm}
    \captionof{figure}{\looseness=-1\textit{Role of minimality}: We plot the DCI metric of a set of models (\textit{red dots}) trained on fixed tasks from \texttt{DSprites}: Training without regularizers leads to no disentanglement (\textit{green}). Enforcing sparsity alone (\textit{yellow}, akin to \cite{lachapelle2022synergies}) achieves good disentanglement ($DCI=71.9$), but some features may be split or duplicated. Enforcing both minimality and sparse sufficiency (\textit{magenta}) attains the best $DCI$ ($98.8$). When $\beta$ is too high ($>0.25$) activated features collapses into few clusters with respect to tasks. For complete results and experiments on additional datasets see Table \ref{tab:betavsdci} and Figures \ref{fig:betavsDCI3DShapes}, \ref{fig:disentanglement_qualitative} in Appendix.}
 \label{fig:betavsDCI}
 \vspace{-0.4cm}
\end{wrapfigure}

We then randomly draw and fix 2 groups of tasks with supports $S_1,S_2$ (18 in total), which all have support on two FOVs, $|S_1|=|S_2|=2$. The groups share one factor of variation and differ in the other one, i.e. $S_1 \cap S_2 =\{i\}$ for some $\{i\} \in Z$. The data in these tasks are subject to spurious correlations, i.e. FOVs not in the task support may be spuriously correlated with the task label. We start from an overestimate of the dimension of $\tilde{\mathbf{z}}$ of $6$, trying to recover $\mathbf{z}$ of size $3$. We train our network to solve these tasks, enforcing sufficiency and minimality on the representation with different regularization degrees. In Figure \ref{fig:betavsDCI}, we show how the alignment of the learned features with the ground truth factors of variations depend on the choice of $\alpha, \beta$, going from no disentanglement ($DCI=27.8$). to good alignment as we enforce more sufficiency and minimality. 
The model that attains the best alignment ($DCI=98.8$) uses both sparsity and feature sharing.  Sufficiency alone (akin to the method of \cite{lachapelle2022synergies}) is able to select the right support for each task, but features are split or duplicated, attaining lower disentanglement ($DCI=71.9$). The feature sharing penalty ensures clustering in the feature space w.r.t. tasks, ensuring to reach high disentanglement, although it may result in the failure cases, when $\beta$ is too high ($\beta>0.25$).

\begin{table}[ht!]
    \caption{\textit{Enforcing disentanglement}: DCI \cite{eastwood2018} disentanglement scores and ID accuracy on test samples for a model trained without enforcing sufficiency and minimality (top row), and model with the regularizers activated (bottom row). While attaining similar performance on accuracy, the model with the activated regularizer always show higher disentanglement. See Table \ref{tab:full_results_synthetic} for additional scores.}%\FL{in the text you say 98.8, why is it not here? Should we add Seb's method here?}}\MF{the 98.8 score is referred to figure 3}
    \centering
    \begin{tabular}{ccccc}
    \toprule 
        & Dsprites & 3Dshapes & SmallNorb & Cars\\
    \midrule 
           \textit{No reg}&          &           &   &\\    
      (DCI,Acc) &   (16.6,94.4)       &  (44.4,96.2 )         &    (16.5,96.1)    & (60.5,99.8)\\
    \midrule 
         $\alpha,\beta$  &        &              &        &  \\

      (DCI,Acc)  &   ($\mathbf{69.9}$,95.8)        &  ($\mathbf{87.7}$,  95.8)             &   ($\mathbf{55.8}$,95.6 )      & ($\mathbf{92.3}$,99.8 ) \\

    \bottomrule 
    \end{tabular}
    \label{tab:disentanglement}
\end{table}

% \begin{table}[h!]
%     \centering
%     \resizebox{\linewidth}{!}{
%     \begin{tabular}{|c|c|c|c|}
%     \hline 
%        \small \textit{(DCI,ACC)} & Dsprites & 3Dshapes & SmallNorb \\
%     \hline 
%       $\cancel{\alpha},\cancel{\beta}$&   16.6/94.4        &  44.4/96.2            &    16.5/96.1    \\
%     \hline 
%      $\alpha,\beta$  &   $\mathbf{69.9}$/95.8        &  $\mathbf{87.7}$/97.8              &   $\mathbf{55.8}$/95.6        \\

%     \hline 
%     \end{tabular}}
%     \caption{DCI scores vs accuracy on test.}
%     \label{tab:disentanglement}
% \end{table}

\begin{wrapfigure}{R}{0.5\linewidth}
    \includegraphics[width=\linewidth,trim={0.5cm 0 0.9cm 1.1cm},clip]{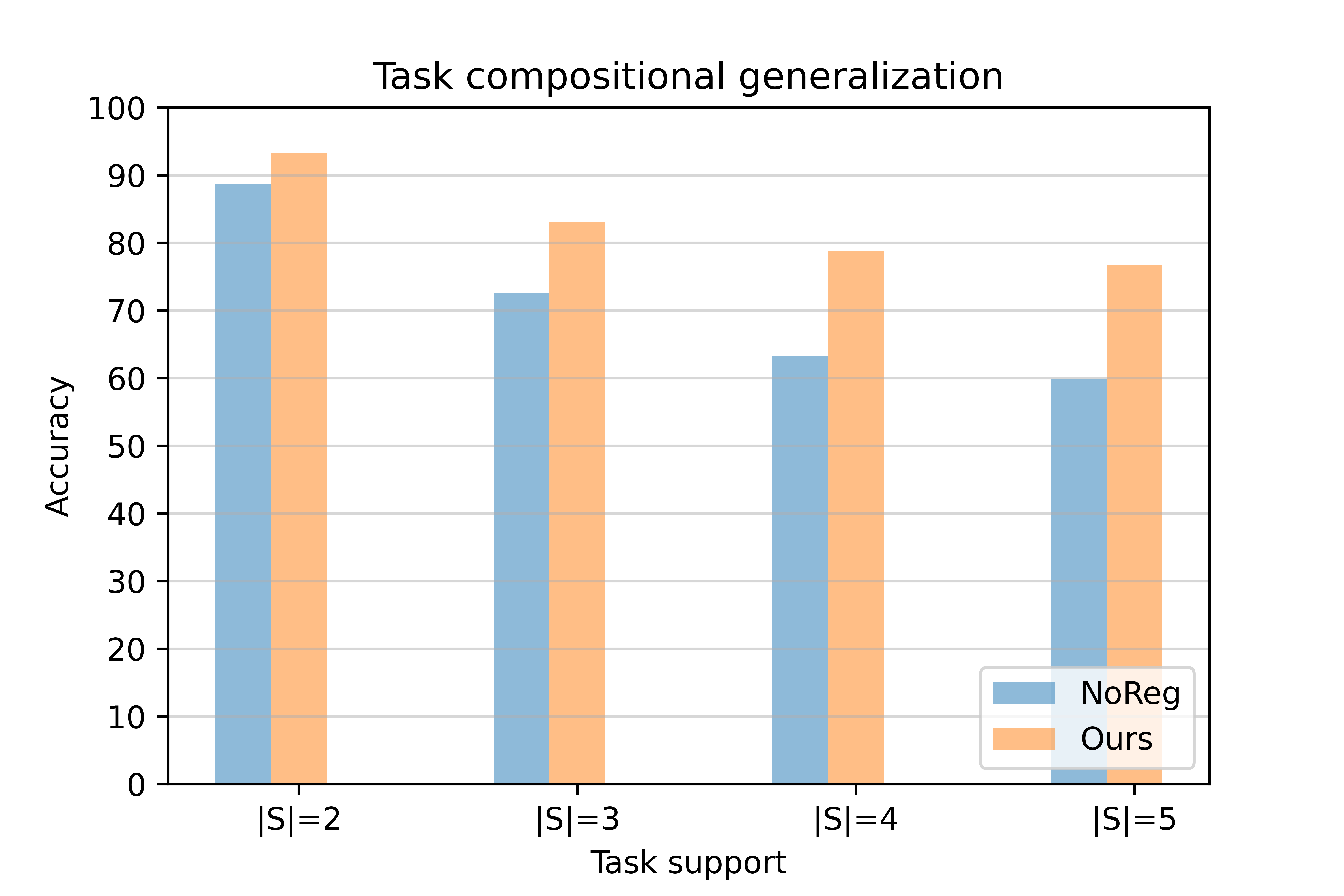}
     \vspace{-0.6cm}
    \captionof{figure}{\emph{Task compositional generalization}: Mean accuracy over 100 
 random test tasks reported for group of tasks of growing support (\textit{second, third, fourth column}) for a model trained without inductive biases (\textit{blue}, attaining $DCI=29.4$) and enforcing them (\textit{orange}, $DCI=59.4$). The latter show better compositional generalization resulting from the properties enforced on the representation. Exact values are reported in Table \ref{tab:task_generalization} in Appendix.}
 \label{fig:task_composition}
 \vspace{-0.5cm}
\end{wrapfigure}

\textbf{Disentanglement and minimality are correlated.} \label{sec:corr_dis_beta}
In the synthetic setting, we also show the role of the feature sharing penalty. Minimizing the entropy of feature activations across mini-batches of tasks results in clusters in the feature space. %In Figure \ref{fig:disentanglement_quantitative}
We investigate how the strength of this penalty correlates well with  disentanglement metrics \cite{eastwood2018} training different models on \texttt{Dsprites} which differ by the value of $\beta$. For 15 models trained increasing $\beta$ from $0$ to $0.2$ linearly, we observe a correlation coefficient with the DCI metric associated to representations compute by each model of $94.7$, showing that  the feature sharing property strongly encourages disentanglement. This confirms again that sufficiency alone (i.e. enforcing sparsity) is not enough to attain good disentanglement.

\textbf{Task compositional generalization.} 
Finally, we evaluate the generalization capabilities of the features learned by our method by testing our model on a set of unseen tasks obtained by combining tasks seen during training. To do this, we first train two models on the \texttt{AbstractDSprites} dataset using a random distribution of tasks, where we limit the support of each task to be within 2 (i.e. $|S|=2$). The models differ in  activating/deactivating the regularizers on the linear heads. Then, we test on 100 tasks drawn from a distribution with increasing support on the factors of variation $(|S|=3,|S|=4,|S|=5)$, which correspond to composition of tasks in the training distribution; see Figure \ref{fig:task_composition}, with the accompaning Table~\ref{tab:task_generalization} in Appendix \ref{sec:app_additional_results}.

% \begin{figure}[h!]
% \centering
% \resizebox{0.8\linewidth}{!}{
% \begin{tabular}{l}
% \includegraphics[trim={0 0 0 1.1cm},width=0.5\linewidth,clip]{pictures/DSprites_Nep_300_DCI.png} %& \includegraphics[trim={0 0 0 1.1cm},width=0.5\linewidth,clip]{pictures/DSprites_Nep_300_MIG.png}
% \end{tabular}}
% \caption{\textit{Minimality encourages disentanglement}: Plots showing correlations between strenghts of the feature sharing penalty $\beta$ (x axis) and the DCI disentanglement measures (y axis)}.% \emph{Left} DCI measure. \emph{Right} MIG measure.}
% \label{fig:disentanglement_quantitative}
% \end{figure}

\subsection{Domain Generalization} \label{sec:dg}

\begin{table}%[h!]
\caption{Quantitative results for few-shot transfer learning, with our method consistently outperforming ERM across all sample sizes and data sets.}
\centering
\begin{tabular}{ccccc}
\toprule
\multicolumn{2}{l}{\textbf{N-shot/Algorithm}} & \multicolumn{3}{c}{\textbf{OOD accuracy (averaged by domains)}} \\
\midrule
{\textbf{1-shot}} &         PACS &           VLCS &   OfficeHome &        Waterbirds \\

ERM       &  80.5  &      $59.7$    &  56.4  & 79.8  \\
Ours       &  $\mathbf{81.5}$  &      $\mathbf{68.2}$ &  $\mathbf{58.4}$   &  $\mathbf{88.4}$  \\
\midrule

{\textbf{5-shot}} &          & &    &         \\

ERM       & 87.1  &      71.7    &  75.7  & 79.8   \\
Ours       & $\mathbf{88.3}$ &      $\mathbf{74.5}$ &  $\mathbf{77.0}$    &  $\mathbf{87.6}$    \\
\midrule

{\textbf{10-shot}} &         &            &    &         \\

ERM       & 87.9  &      74.0    &  81.0   & 84.2   \\
Ours      & $\mathbf{90.4}$  &      $\mathbf{77.3}$   &  $\mathbf{82.0}$   &  $\mathbf{89.2}$   \\
\bottomrule

\end{tabular}

    \vspace{-1cm}
\label{tab:fs_transfer}
\end{table}

In this section we evaluate our method on benchmarks coming from the domain generalization field~\cite{gulrajani2020search, Wenzel2022AssayingOG, multimodal_robustness} and subpopulation distribution shifts \cite{Sagawa2019,koh2021wilds}, to show that a feature space learned with our inductive biases performs well out of real world data distribution. % , proving uselfulness of disentangled representations for real settings, where the factors of variations are unknown.  

\begin{wrapfigure}{R}{0.45\linewidth}
\vspace{-0.6cm}
\centering
\includegraphics[width=\linewidth,trim={0.1cm 0cm 0.9cm 1.1cm},clip]{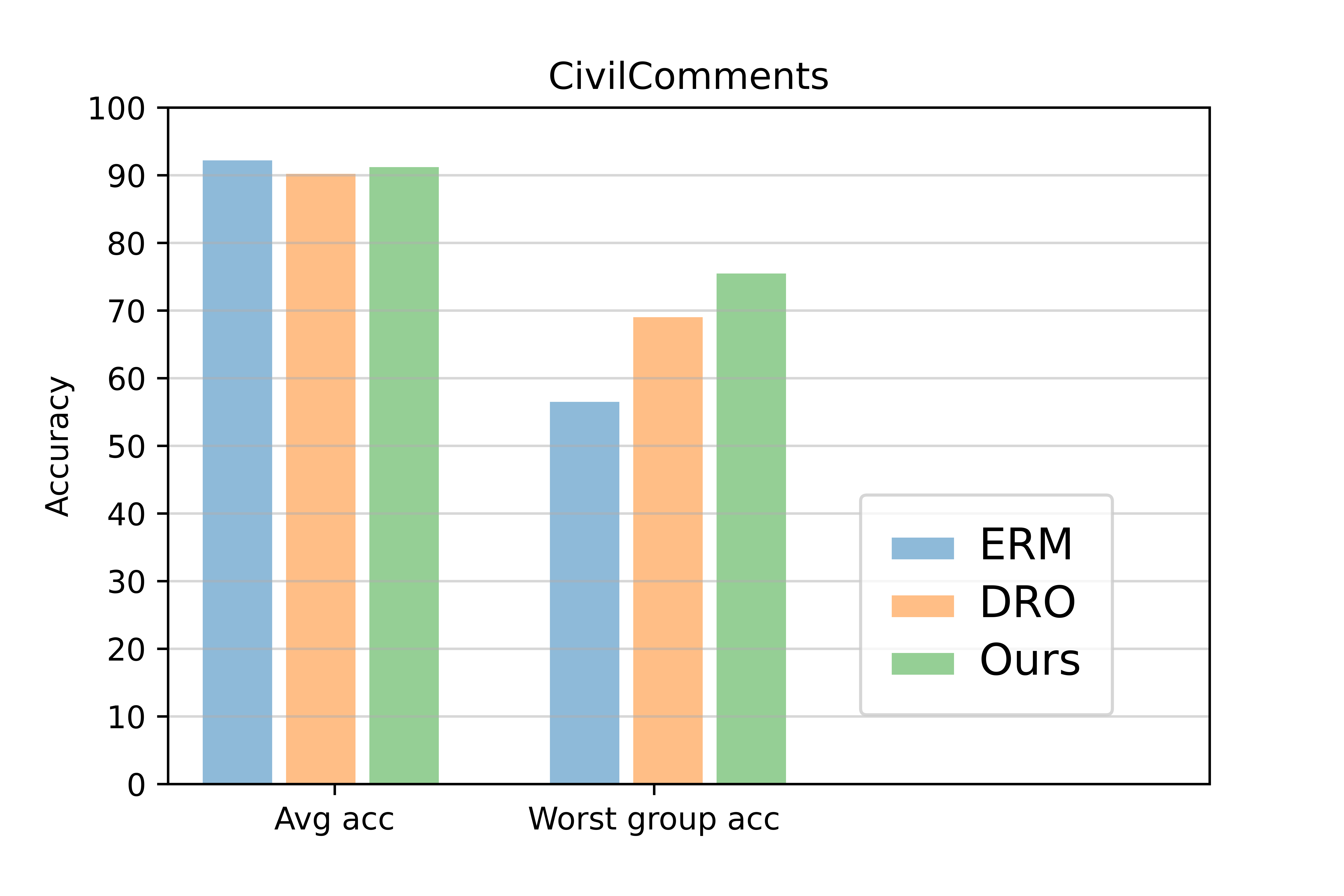}
\vspace{-0.6cm}
\captionof{figure}{\textit{Quantitative results on CivilComments}: we report the accuracy on test averaged across all demographic groups (\textit{left group}), and the worst group accuracy, on the \textit{right}. Our method (\textit{green}) performs similarly in terms of average accuracy and outperforms in terms of worst group accuracy, without using any knowledge on the group composition in the training data. For exact values and error estimates, see Table \ref{tab:civilcomments} in the Appendix.}
\label{fig:civilcomments}
\vspace{-0.7cm}
\end{wrapfigure}

\textbf{Subpopulation shifts.} \
Subpopulation shifts occur when the distribution of minority groups changes across domains. Our claim is that a feature space that satisfies sparse sufficiency and minimality is more robust to spurious correlations which may affect minority groups, and should transfer better to new distributions. To validate this, we test on two benchmarks \texttt{Waterbirds} \cite{Sagawa2019}, and \texttt{CivilComments}~\cite{koh2021wilds} (see Appendix \ref{sec:app_dataset_description}).

For both, we use the train and test split of the original dataset.
In Table \ref{tab:dg_domainbed}, last row, we report the results on the test set of \texttt{Waterbirds} for the different groups in the dataset (landbirds on land, landbirds on water, waterbirds on land, and waterbirds on water, respectively). We fit the linear head on a random subset of the training domain, balanced by class, repeat 10 times and report accuracy and standard deviation on test. For \texttt{CivilComments} we report the average and worst accuracy in Figure~\ref{fig:civilcomments}, where we compare with ERM and groupDRO \cite{Sagawa2019}. While performing almost on par w.r.t. ERM, our method is more robust to spurious correlation in the dataset, showing the higher worst group accuracy. Importantly, we outperform GroupDRO, which uses information on the subdomain statistics, while we do not assume any prior knowledge about them. Results per group are reported in the Appendix (Table~\ref{tab:civilcommentsgroups}).

% \begin{table}[]
%     \caption{Quantitative results on CivilComments: we report the accuracy on test averaged across all demographic groups (\textit{left}), and the worst group accuracy (\textit{right}). We show that our method performs similarly in terms of average accuracy and outperforms in terms of worst group accuracy, without using any knowledge on the group composition in the training data.}
%     \label{tab:civilcomments}
%     \centering
%     \begin{tabular}{ccc}
%     \toprule 
%       & avg acc  & worst group  acc\\
%     \midrule 
%       ERM  & $\mathbf{92.2}$ & 56.5   \\
%       DRO  &  90.2 & 69  \\
%       Ours &  91.2 \footnotesize$\pm$ 0.2 & $\mathbf{75.45}$\footnotesize$\pm$  0.1\\
%         \bottomrule 
%     \end{tabular}
% \end{table}

%In Figure \ref{fig:waterbirds_quantitative} we shot the accuracy scores of our model compared to an ERM baseline n a 1-shot, 5-shot and 10-shot setting.

% \begin{figure}[h]
% \centering
% \resizebox{\linewidth}{!}{
% \begin{tabular}{lll}
%  \includegraphics[width=0.5\linewidth]{pictures/waterbirds1-shot.png}
% & \includegraphics[width=0.5\linewidth]{pictures/5shotwaterbirds.png} &  \includegraphics[width=0.5\linewidth]{pictures/10shotwaterbirds.png}
% \end{tabular}}
% \caption{Waterbirds quantitative: ours vs ERM}
% \label{fig:waterbirds_quantitative}
% \end{figure}
\textbf{DomainBed.} We evaluate the domain generalization performance on the \texttt{PACS}, \texttt{VLCS} and \texttt{OfficeHome} datasets from the DomainBed \cite{gulrajani2020search} test suite (see Appendix \ref{sec:app_dataset_description} for more details). % It contains 15,500 images, with an average of around 70 images per class and a maximum of 99 images in a class.
On these datasets, we train on $N-1$ and leave one out for testing. Regularization parameters $\alpha$ and $\beta$ are tuned according to validation sets of \texttt{PACS}, and used accordingly on the other dataset. For these experiments we use a \texttt{ResNet50} pretrained on \texttt{Imagenet}  \cite{imagenet_cvpr09} as a backbone, as done in \cite{gulrajani2020search} %\FL{Say this is standard for the benchmark}.
%\FL{Is this standard?}\MF{it's not, cause with other methods you usually use directly the head that you have trained. For us that is task specific, so we need to fit another one at test time. There was anyway a work \cite{kirichenko2022last} where they fit a linear head  sampling data points from the training set in balanced way with respect to subgroups to be robust on distribution shift tasks. With respect to them we dont assume to know the subgroups or the spurious correlations, so we just do different samples and report the std}.
To fit the linear head we sample 10 times with different samples sizes from the training domains and we report the mean score and standard deviation. Results are reported in Table \ref{tab:dg_domainbed}, showing how enforcing sparse sufficiency and minimality leads consistently to better OOD performance. Comparisons with 13 additional baselines is in Appendix~\ref{sec:app_full_Dg_exp}.
\begin{table}%[ht!]
    \caption{Quantitative evaluation on Camelyon17: we report accuracy both on ID and OOD splits. Our approach achieves significantly higher validation and test OOD accuracy.}
    \label{tab:camelyon}
    \centering
    \begin{tabular}{cccc}
    \toprule 
         & Validation(ID) & Validation (OOD) & Test (OOD) \\
    \midrule 
     ERM &  93.2          & 84               &  70.3      \\
     CORAL &  $\mathbf{95.4}$     & 86.2             &  59.5     \\
     IRM &  91.6          & 86.2             &  64.2     \\
     Ours &  93.2 \footnotesize±0.3         & $\mathbf{89.9}$\footnotesize±0.6         &  $\mathbf{74.1}$\footnotesize±0.2     \\
    \bottomrule 
    \end{tabular}
    \vspace{-1cm}
\end{table}

\textbf{Camelyon17.}
The model is trained according to the original splits in the dataset. In Table \ref{tab:camelyon} we report the accuracy of our model on in-distribution and OOD splits, compared with different baselines \cite{sun2017,arjovsky2019}. Our method shows the best performance on the OOD test domains. The intuition of why this happens is that, due to minimality, we retain more features which are shared across the three training domains, giving less importance to the ones that are domain-specific (which contain the spurious correlations with the hospital environmental informations). This can be further enforced at test time, as we show in the ablation in Appendix \ref{sec:app_feature_sharing_test_time}, trading off in distribution performance for OOD accuracy. %\FL{this is super confusing. Let's remove this and make it an ablation (maybe even in the appendix)} /MF{Done}
%\MF{Are we sure to remove the following?}Importantly, for this dataset, we demonstrate the benefits of utilizing the feature sharing penalty at test time. The last row of Table~\ref{tab:camelyon} illustrates how the performance changes when optimizing feature sharing at test time. The results show that performance slightly decreases on the training and validation domains, but significantly improves on the test domain, whose distribution is different from the others. The intuition is that we are retaining features which are shared across the three training domains and cutting the ones that are domain-specific (which contains the spurious correlations with the hospital environment). 

% \begin{table}[h!]
% \centering
% \resizebox{\linewidth}{!}{
% \begin{tabular}{|c|c|c|c|c|c|}
% \hline
% \multicolumn{2}{l}{\textbf{Dataset/Algorithm}} & \multicolumn{3}{c}{\textbf{OOD accuracy (by domain)}} \\
% \hline
% {\textbf{Waterbirds}} &         LL &           LW &   WL &        WW & Average\\

% ERM       &  98.6 \footnotesize$\pm$ 0.3 &       52.05 \footnotesize$\pm$  3 &  68.5 \footnotesize$\pm$ 3 & 93 \footnotesize$\pm$ 0.3 & 81.3\\
% Ours       &  $\mathbf{99.5}$ \footnotesize$\pm$ 0.1&      $\mathbf{73.0}$ \footnotesize$\pm$ 2.5 &  $\mathbf{85.0}$ \footnotesize$\pm$ 2  &  95.5 \footnotesize$\pm$ 0.4 & $\mathbf{90.5}$ \\
% \hline

% \end{tabular}
% }
% \caption{Results on Waterbirds}
% \label{tab:dg_waterbirds}
% \end{table}

\begin{table}%[ht!]
\caption{Results for domain generalization on DomainBed. Our approach achieves consistently higher average OOD generalization, outperforming ERM in all cases except one.}
\label{tab:dg_domainbed}
\centering
\begin{tabular}{cccccc}
\toprule
\multicolumn{2}{l}{\textbf{Dataset/Algorithm}} & \multicolumn{3}{c}{\textbf{OOD accuracy (by domain)}} \\
\midrule
{\textbf{PACS}} &         S &           A &   P &        C & Average\\

ERM       &  77.9 \footnotesize$\pm$ 0.4 &      $\mathbf{88.1}$ \footnotesize$\pm$  0.1 &  97.8 \footnotesize$\pm$ 0.0 & 79.1 \footnotesize$\pm$ 0.9 & 85.7\\
Ours       &  $\mathbf{83.1}$ \footnotesize$\pm$ 0.1&      86.7\footnotesize$\pm$ 0.8 &  $\mathbf{97.8}$ \footnotesize$\pm$ 0.1  &  $\mathbf{83.5}$ \footnotesize$\pm$ 0.1 & $\mathbf{87.5}$ \\
\midrule
{\textbf{VLCS}} &         C &           L &   V &        S & Average\\
ERM       &  97.6\footnotesize$\pm$ 1.0 &      63.3 \footnotesize$\pm$ 0.9  & 76.4 \footnotesize$\pm$ 1.5 &   72.2 \footnotesize$\pm$  0.5 & 77.4\\
Ours       &  $\mathbf{98.1}$\footnotesize$\pm$ 0.2 &      $\mathbf{63.4}$\footnotesize$\pm$ 0.5 & $\mathbf{78.2}$ \footnotesize$\pm$  0.7 &  $\mathbf{73.9}$\footnotesize$\pm$ 0.8 & $\mathbf{78.4}$ \\
\midrule
{\textbf{OfficeHome}} &         C &           A &   P &        R & Average\\
ERM       &  53.4\footnotesize$\pm$ 0.6 &      62.7 \footnotesize$\pm$ 1.1 & 76.5 \footnotesize$\pm$ 0.4 &  77.3 \footnotesize$\pm$ 0. & 67.5\\
Ours       &  $\mathbf{56.3}$\footnotesize $\pm$ 0.1 &      $\mathbf{66.7}$ \footnotesize$\pm$ 0.7 & $\mathbf{79.2}$\footnotesize$\pm$ 0.5 &  $\mathbf{81.3}$ \footnotesize$\pm$ 0.4 &  $\mathbf{70.9}$\\

\midrule
{\textbf{Waterbirds}} &         LL &           LW &   WL &        WW & Average\\

ERM       &  98.6 \footnotesize$\pm$ 0.3 &       52.05 \footnotesize$\pm$  3 &  68.5 \footnotesize$\pm$ 3 & 93 \footnotesize$\pm$ 0.3 & 81.3\\
Ours       &  $\mathbf{99.5}$ \footnotesize$\pm$ 0.1&      $\mathbf{73.0}$ \footnotesize$\pm$ 2.5 &  $\mathbf{85.0}$ \footnotesize$\pm$ 2  &  $\mathbf{95.5}$ \footnotesize$\pm$ 0.4 & $\mathbf{90.5}$ \\
\bottomrule
    \vspace{-1cm}

\end{tabular}
\end{table}

\subsection{Few-shot transfer learning.}
We finally show the ability of features learned with our method to adapt to a new domain with a small number of samples in a few-shot setting. 
We compare the results with ERM in Table \ref{tab:fs_transfer}, averaged by domains in each benchmark dataset. The full scores for each domain are in Appendix \ref{sec:app_fewshot} for 1-shot, 5-shot, and 10-shot setting, reporting the mean accuracy and standard deviations over 100 draws. Our approach achieves consistently higher accuracy than ERM, showing the better adaptation capabilities of our minimal and sufficently sparse feature space.
\subsection{Additional results}
\looseness=-1In Appendix \ref{sec:app_additional_results} we report a large collection of additional results, including comparison with 14 baseline methods on the domain shift benchmarks 
 (\ref{sec:app_full_Dg_exp}), a qualitative and quantitative analysis on the minimality and sparse sufficiency properties in the real setting (\ref{sec:app_properties_analysis}), a favorable additional comparison on meta learning benchmarks, with 6 other baselines including ~\cite{lachapelle2022synergies}(\ref{sec:app_meta_learning}), an ablation study on the effect of clustering features at test time  (\ref{sec:app_feature_sharing_test_time}), and a demonstration on the possibility to obtain a task similarity measure as a consequence of our approach (\ref{sec:app_task_sim}).

\section{Conclusions}
\looseness=-1In this paper, we demonstrated how to learn disentangled representations from a distribution of tasks by enforcing feature sparsity and sharing. We have shown this setting is identifiable and have validated it experimentally in a synthetic and controlled setting. Additionally, we have empirically shown that these representations are beneficial for generalizing out-of-distribution in real-world settings, isolating spurious and domain specific factors that should not be used under distribution shift.

\textbf{Limitations and future work}: The main limitation of our work is the global assumption on the strength of the sparsity and feature sharing regularizers $\alpha$ and $\beta$ across all tasks. In real settings these properties of the representations might need to change for different tasks.  
We have already observed this in the synthetic setting in Figure~\ref{fig:betavsDCI}, where when $\beta>0.25$ features cluster excessively and are unable to achieve clear disentanglement and do not generalize well. 
Future work may exploit some level of knowledge on the task distribution (e.g. some measure of distance on tasks) in order to tune $\alpha, \beta$ adaptively during training, or to train conditioning on a distribution of regularization parameters as in \cite{dosovitskiy2020you}, enabling more generalization at test time.
Another limitation is in the sampling procedure to fit the linear head at test time: sampling randomly from the training set (balanced by class) may not  be enough to achieve the best performance under distributions shifts. Alternative sampling procedures, e.g. ones that incorporate knowledge on the distribution shift if available (as in \cite{kirichenko2022last}), may lead to better performance at test time.

\newpage
\begin{ack}
Marco Fumero and Emanuele Rodolà were supported by the ERC grant no.802554 (SPECGEO), PRIN 2020 project 
no.2020TA3K9N (LEGO.AI), and PNRR MUR project PE0000013-FAIR. Marco Fumero and Francesco Locatello were partially at Amazon while working at this project. We thank Julius von Kügelgen, Sebastian Lachapelle and the anonymous reviewers for their feedback and suggestions.
\end{ack}

\bibliographystyle{plain}
\bibliography{main}

%%%%%%%%%%%%%%%%%%%%%%%%%%%%%%%%%%%%%%%%%%%%%%%%%%%%%%%%%%%%%%%%%%%%%%%%%%%%%%%
%%%%%%%%%%%%%%%%%%%%%%%%%%%%%%%%%%%%%%%%%%%%%%%%%%%%%%%%%%%%%%%%%%%%%%%%%%%%%%%
% APPENDIX
%%%%%%%%%%%%%%%%%%%%%%%%%%%%%%%%%%%%%%%%%%%%%%%%%%%%%%%%%%%%%%%%%%%%%%%%%%%%%%%
%%%%%%%%%%%%%%%%%%%%%%%%%%%%%%%%%%%%%%%%%%%%%%%%%%%%%%%%%%%%%%%%%%%%%%%%%%%%%%%
\newpage
\appendix
\section{Proof of Proposition 1}
To prove Proposition \ref{thm:ident} we rely on the same proof construction of \cite{locatello2020weakly}, adapting it to our setting. Intuitively, the proposition states that when minimality and sparse sufficiency properties hold it is possible to recover the factors of variations $z$ given enough observations from $p(x,y)$, if the following assumptions on the task distribution hold: (i) the probability of two arbitrary tasks having a singleton intersection of support on the factor of variations is non zero; (ii) the probability that their difference of supports is a singleton is non zero.

The proof is sketched in three steps:
\begin{itemize}
    \item First, we prove identifiability when the support $S$ of a task is arbitrary but fixed, where we drop the subscript $t$ for convenience.
    \item Second, we randomize on $S$, to extend the proof for $S$ drawn at random.
    \item Third, we extend the proof to the case when the dimensionality of $\mathcal{Z}$ is unknown and we start on overestimate of it to recover it.
\end{itemize}

\textbf{Identifiability with fixed task support}
We assume the existence of the generative model in Figure \ref{fig:causal_model}, which we report here for convenience:
\begin{align}p(\mathbf{z})=&\prod_ip(z_i) \ \  \ \ \ \ \ \ \ \ \ \ \ \ \ \ \   &&S \sim p(S) \\ &\mathbf{x} = g^*(\mathbf{z}) && y=f_S^*(\mathbf{z})
\end{align} 
together with the assumptions specified in theorem statement. 
We fix the support of the task $S$.
We indicate with $g: Z \rightarrow X$ the invertible smooth, candidate function we are going to consider, whose inverse corresponds to $q(\mathbf{z}|\mathbf{x})$.  We denote with $T \in S$ which indexes the coordinate subspace of image of $g^{-1}$ corresponding to the unknown coordinate subspace $S$ of factors of variation on which the fixed task depends on. Fixing $T$ requires knowledge of $|S|$.
The candidate function $g^{-1}$ must satisfy: 
\begin{align}
f|_T(g^{-1}(\mathbf{x}))=y \\f|_{\bar{T}}(g^{-1}(\mathbf{x}))\neq y\end{align}

where $\bar{T}$ denotes the indices in the complement of $T$. $f$ denotes a predictor which satisfies the same assumptions on $f^*$ on $T$ .
We parametrize $g^{-1}$ with $g^{*-1}$ and set:

$g^{-1}=h^{-1}\circ g^{*-1}$ where $h:[0,1]^d \rightarrow Z$, mapping from the uniform distribution on $\mathbb{R}^d$ to $Z$.
We can rewrite the two above constraints as:

\begin{align}f|_T(h^{-1}(z))=y \\ f|_{\bar{T}}(h^{-1}(z)) \neq y\end{align}
We claim that the only admissible functions $h^{-1}$ maps each entry in $\mathbf{z}$ to unique coordinate in $T$.
We observe that due to its smoothness and invertibility, $h^{-1}$ maps $Z$ to the submanifolds $\mathcal{M}_s,\mathcal{M}_{\bar s}$, which are disjoint.
By contradiction:
\begin{itemize}
    \item if $\mathcal{M}_{\bar{S}}$ does not lie in $\bar{T}$ then minimality is violated.
    \item if $\mathcal{M}_S$ does not lie in $T$ then sufficiency is violated
\end{itemize}

$h^{-1}$ maps each entry in $\mathbf{z}$ to unique coordinate in $T$. Therefore there exist a permutation $\pi$ s.t.: 
\begin{align}h_T^{-1}(\mathbf{z})=\bar{h}_T(\mathbf{z}_{\pi(S)}) \\ h_{\bar{T}}^{-1}(\mathbf{z})=\bar{h}_{\bar{T}}(\mathbf{z}_{\pi(\bar{S})})\end{align}
The Jacobian of $h^{-1}$  is  a blockwise matrix with block indexed by $T$. So we can identify the two blocks of factors in $S, \bar{S}$ but not necessarily the factors within, as they may be still entangled.
\newpage
\textbf{Randomization on $S$}

we now consider $S$ to be drawn at random, therefore we observe $p(\mathbf{x},y |S)$ without never observing $S$ directly. $g^{-1}$ must now associate each $p(\mathbf{x},y)$ with a unique $T$, as well as a unique predictor $f$ , for each $S \sim p(S)$
Indeed suppose that $p(\mathbf{x},y |S=S_1)$ and $p(\mathbf{x},y |S=S_2)$ with $S_1,S_2 \sim p(S)$ and  $S_1 \neq S_2$.
Then if $T$  would be the same for both tasks (as $f$), eq (6) could only be satisfied for a subset of size $|S_1 \cap S_2 | <|S_1 \cup S_2|$ , while $T$ is required to be of size  $|S_1 \cup S_2|$
This corresponds to say that each task has its own sparse support and its own predictor.
Conversely all $p(\mathbf{x},y) \in supp(p(\mathbf{x},y |S))$  need to be associated to the $T$ and the same predictor $f$, since they will all share the same subspace and cannot be associated to different $T$.
Notice also that $|S_1 \cap S_2| =|T_1 \cap T_2 |$ and $|S_1 \cup S_2| =|T_1 \cup T_2 |$.
We further assume: 

$\forall z_i$  either $p(S \cap S' = \{i\})> 0$ or  $p( \{i\} \in (S \cup S') -(S'\cap S))>0$

We observe every factor as the intersection of the sets $S,S'$  which will be reflected  in $T,T'$ or we observe single factors in the  difference between the intersection and the union of $S,S'$. Examples of the two cases are illustrated below:
\begin{table}[h]
    \centering
    \begin{tabular}{cc}
       \includegraphics[width=0.3\linewidth]{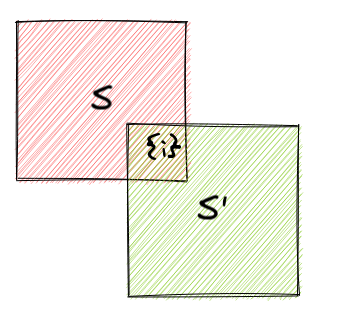}  &  \includegraphics[width=0.3\linewidth]{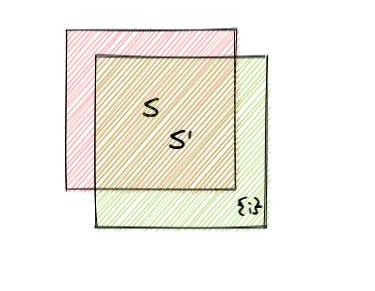}
    \end{tabular}
\end{table}

This together with (8) and (9) implies:

\begin{align} h_i^{-1}(\mathbf{z}) = \bar{h}_i(z_{\pi(i)}) \ \ \ \  \forall i 
\in [d] \end{align}

This further implies that the jacobian of $\bar{h}$ is diagonal. By the change of variable formula we have:

\begin{align}q(\mathbf{\hat{z}}) =p(\tilde{h}(\mathbf{z}_{\pi( [d])})) \left| det\frac{\partial}{\partial\mathbf{z}_{\pi([d]))}}\tilde{h} \right| = \prod_{i01}^d p(\tilde{h_i}(z_{\pi(i)})) \left| \frac{\partial}{\partial z_{\pi(i)}} \tilde{h}_i \right|\end{align}

This holds for the jacobian being diagonal and invertibility of $\tilde{h}$. Therefore $q(\hat{\mathbf{z}})$ is a coordinate-wise reparametrization of $p(\mathbf{z})$ up to a permutation of the indices. A change in a coordinate of $\mathbf{z}$ implies a change in the unique corresponding coordinate of $\hat{\mathbf{z}}$, so $g$ disentangles the factors of variation.

\textbf{Dimensionality of the support $S$}

Previously we assumed that the dimension of $\hat{\mathbf{z}}$ is the same as $\mathbf{z}$. We demonstrate that even when $d$ is unknown starting from an overstimate of it, we can still recover the factors of variations. Specifically, we consider the case when $\hat{d} > d$.
In this case  our assumption  about the invertibility of $h$  is violated. We must instead ensure that $h$ maps $Z$ to a subspace of $\hat{Z}$ with dimension $d$.
To substitute our assumption on inveribility on $h$, we will instead assume that $\mathbf{z}$ and $\hat{\mathbf{z}}$ have the same mutual information with respect to task labels $Y$, i.e.$I (Z,Y)=I (\hat{Z},Y)$
Note that mutual information is invariant to invertible transformation, so this property was also valid  in our previous assumption. 

Now, consider two arbitrary tasks with $|S \cap S'| \neq \emptyset$ =$k$ but $|T \cap T'| <k$, i.e. some features are duplicated/splitted. Hence $f,f'$ while have different support , i.e.:
\begin{align*}
    f|_T=f'|_{T'}=f^*
\end{align*}

\begin{figure}[h]
    \centering
    \includegraphics [width=0.5\linewidth]{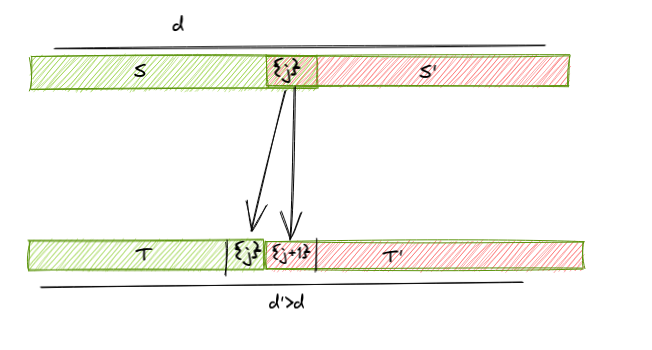}
\end{figure}
We observe that in this situation nor sufficiency, nor minimality are  necessarily violated because:
\begin{itemize}
    \item  $f|_T=f'|_{T'}=f^*$ (sufficiency is not violated)
    \item  $T \cap T'= \emptyset \implies T \not \subset T',T' \not \subset T$ (minimality is not violated)
\end{itemize}
In other words we must ensure that a single fov  $z_i$ is not mapped to different entries in $\hat{\mathbf{z}}$ (feature splitting or duplication).
We fix two arbitrary tasks with $|S \cap S'| \neq \emptyset$ =$k$ but $|T \cap T'| <k$, i.e. some features are duplicated.
We know that $|S|=|T|$ and  $|S'|=|T'|$ otherwise sufficency and minimaliy would be violated.
Then if  $|T \cap T'| <k$, then  $|T \cup T' |>|S \cup S' |=d-k$
we have $p(|T \cup T'|)$=$p(supp(p(y|\hat{\mathbf{z}}))+supp(p'(y'|\hat{\mathbf{z'}})))= p(\sum_isupp(f_i(.))$
 , and since
\begin{align}H [p(\sum_isupp(f_i(.))] >H [p(\sum_isupp(f_i(.))] \end{align}
but  we have assumed:
\begin{align}
I (Z,Y)&=I (\hat{Z},Y) \\
\cancel{H(Y)} -H(Y|\hat{Z})&= \cancel{H(Y)}-H(Y|Z) \\
H(Y|\hat{Z})&= H(Y|Z) \\
H[p(Y|\hat{Z})>0 ]&= H[p(Y|Z)>0 ] \\
2^{H[p(Y|\hat{Z})>0 ]}&=2^{ H[p(Y|Z)>0 ]} \\
|supp(p(Y|\hat{Z}))|&= |supp(p(Y|Z)|
\end{align}
this last passage is due to relation between cardinality and entropy: for uniform distributions the exponential of the entropy is equal to the cardinality of the support of the distribution.
\begin{align}
|supp(f)|= |supp(f^*)|
\end{align}
We know that (12) must hold for every task, therefore:
$\sum_iI (Z,Y_i)=\sum_iI (\hat{Z},Y_i)$ for each $i$
then:
$\sum_i|supp(\hat{f_i})|=\sum_i |supp(f_i^*)|$
$|\bigcup_iT_i|=|\bigcup_iS_i|$
therefore (12) contradicts our assumption (13).

\section{Implementation details}

\subsection{Training algorithm}
\label{sec:app_algo}

\begin{algorithm}[H]
\caption{Training algorithm}
\begin{algorithmic}[1]
\STATE \textbf{Input:} A task distribution  $\mathcal{T}$ 
\WHILE{Not converged}
\STATE  Sample a batch $B_T$ of $T$ tasks $t \sim \mathcal{T}$
\STATE Sample $(U_t,Q_t)$ from each task in the batch
\STATE {\color{gray}{\textbf{$\#$ Inner loop}}}
\FOR{each $t$ in $B_T$}
\STATE Compute $\mathbf{z}_t^U=g_\theta(\mathbf{x}_t^U)$
\ENDFOR
\STATE Solve $\phi^*=  argmin_\phi \frac{1}{T}\sum_t \mathcal{L}_{inner} ( f_\phi(\mathbf{z}_t^U),y_t^U) + Reg(\phi)$
\STATE {\color{gray}{\textbf{$\#$ Outer loop}}}
\FOR{each $t$}
\STATE Compute $\mathbf{z}_t^Q=g_\theta(\mathbf{x}_t^Q)$
\ENDFOR
\STATE Compute $\mathcal{L}_{outer}(f_{\phi^*}(g_\theta (\mathbf{x}_t^Q),y_t^Q))$
\STATE Compute $\frac{\partial \mathcal{L}_{outer}(\theta)}{\partial \theta}$ as in \cite{geng2022training}
\STATE Update $\theta$
\ENDWHILE
\end{algorithmic}
\end{algorithm}

\subsection{Implicit gradients}
In the backward pass, denoting with $\mathcal{L}_{outer}^*=\mathcal{L}_{outer}(f_{\phi}^*(g_\theta(x^Q)),Y^Q)$ denoting the loss computed with respect to the optimal classifier $f_{\phi}^*$  on the query samples $(x^Q,Y^Q)$, we have to compute the following gradient: 
\begin{align}
    \frac{\partial \mathcal{L}_{outer}^*(\theta)}{\partial \theta}= \frac{\partial \mathcal{L}_{outer}(\theta,\phi^*)}{\partial \theta} + \frac{\mathcal{L}_{outer}(\theta,\phi^*)}{\partial {\phi^*}} \frac{\partial \phi^*}{\partial \theta}
\end{align} 
where %ACϕ∗(zθ)
is the algorithm procedure to solve Eq1, i.e. SGD. While  %L(ACϕ∗(zθ))
is just the gradient of the loss evaluated at the solution of the inner problem and can be computed efficiently with standard automatic backpropagation, %dθdθACϕ∗(zθ)
requires further attention. Since the solution to $C_{\phi^*}$ is implemented via and iterative method (SGD), one strategy would be to compute this gradient would be to backpropagate trough the entire optimization trajectory in the inner loop. This strategy however is computational inefficient for many steps, and can suffer also from vanishing gradient problems.

% \section{Additional related work}

% Multitask learning \cite{caruana1997multitask,zhang2018overview} jointly optimizes models on
%  several related tasks. Hard parameter sharing. Soft parameter sharing.
%  \cite{zhai2019visual,evci2022head2toe}
%  Linear MTlearning, kernel methods, sparse multi task learning 
\section{Experimental details} \label{sec:exp_details}

All experiments were performed on a single gpu NVIDIA RTX 3080Ti and implemented with the Pytorch library \cite{pytorch:2019}.

\subsection{Datasets}
We evaluate our method on a synthetic setting on the following benchmarks: \texttt{DSprites}, \texttt{AbstractDSprites}\cite{dsprites17}, \texttt{3Dshapes} \cite{3dshapes18},\texttt{SmallNorb} \cite{lecun2004learning}, \texttt{Cars3D}\cite{reed2015deep} and the semi-synthetic \texttt{Waterbirds} \cite{Sagawa2019}.

For domain generalization and domain adaptation tasks, we evaluate our method on the \cite{gulrajani2020search} and \cite{koh2021wilds} benchmarks, using the following datasets:
\texttt{PACS}\cite{li2017deeper}, \texttt{VLCS}\cite{albuquerque2019generalizing}, \texttt{OfficeHome}\cite{venkateswara2017deep}
\texttt{Camelyon17}\cite{camelyon}, \texttt{CivilComments} \cite{CivilComments}.

\textbf{Dataset descriptions} \label{sec:app_dataset_description}

The \texttt{Waterbirds} dataset \cite{Sagawa2019} is a synthetic dataset where images are composed of cropping out birds from photos in the \texttt{Caltech-UCSD Birds-200-2011 (CUB)} dataset  \cite{wah2011} and transferring them onto backgrounds from the Places dataset \cite{zhou2017}. The dataset contains a large percentage of training samples ($ \approx \%95$) which are spuriously correlated with the background information.

The \texttt{CivilComments} is a dataset of textual reviews annotated with demographics information for the task of detecting toxic comments. Prior work has
shown that toxicity classifiers can pick up on biases in the training data and spuriously associate toxicity with the mention of certain demographics \cite{park:2018,dixon18}. These types
of spurious correlations can significantly degrade model performance on particular subpopulations \cite{sagawa2020investigation}.

The \texttt{PACS} dataset \cite{li2017deeper} is a collection of images coming from four different domains: \textit{real images, art paintings, cartoon} and  \textit{sketch}.
The \texttt{VLCS}  dataset contains examples from 5 overlapping classes from the VOC2007 \cite{pascal-voc-2007}, LabelMe \cite{russell2008labelme}, Caltech-101 \cite{fei2004learning} , and SUN \cite{xiao2010sun} datasets.
The \texttt{OfficeHome} dataset contains 4 domains (Art, ClipArt, Product, real-world) where each domain consists of 65 categories.

The \texttt{Camelyon17} dataset, is a collection of medical tissue patches scanned from different hospital environments. The task is to predict whether a patch contain a benign or tumoral tissue. The different hospitals represent the different domains in this problem, and the aim is to learn a predictor which is robust to changes in factors of variation across different hospitals.

\subsection{Models}
For synthetic datasets we use a CNN module for the backbone $g\theta$ following the architecture in Table \ref{tab:conv_arch}.
For real datasets that use images as modality we use a \texttt{ResNet50} architecure as backbone pretrained on the \texttt{Imagenet} dataset. For the experiments on the text modality we use \texttt{DistilBERT} model \cite{sanh2019distilbert} with pretrained weights downloaded from HuggingFace \cite{wolf2019huggingface}.

\subsection{Synthetic experiments}

\begin{table}[h!]
    \centering
\caption{Convolutional architecture used in synthetic experiments.}
\label{tab:conv_arch}

    \begin{tabular}{c}
    \toprule 
       \textbf{CNN backbone}\\
    \midrule 
Input : $64 \times 64 \times$ number of channels \\
$4 \times 4$conv, $32$  stride $2$, padding $1$, ReLU,BN \\
$4 \times 4$conv, $32$  stride $2$, padding $1$, ReLU,BN\\
$4 \times 4$conv, $64$  stride $2$, padding $1$, ReLU,BN\\
$4 \times 4$conv, $64$  stride $2$, padding $1$, ReLU,BN\\
FC, $256$, Tanh\\
FC, $d$  \\
        \bottomrule 
\end{tabular}
\end{table}

\textbf{Task generation}. \label{sec:app_task_gen}
For the synthetic experiments we have access to the ground truth factors of variations $\mathcal{Z}$ for each dataset.
The task generation procedure relies on two hyperparameters: the first one is an index set $\mathbb{S}$ of possible factors of variations on which the distribution of tasks can depend on. The latter hyperparameter $K$, set the maximum number of factors of variations on which a single task can depend on.
Then a task $t$ is sampled drawing a number $k_t$ from $\{1...K\}$, and then sampling randomly a subset $S$ of size $|\mathbb{S}| -k_t$ from $\mathbb{S}$. The resulting set $S$ will be the set indexing the factors of variation in Z on which the task $t$ is defined.
In this setting restrict ourselves to binary task: for each factors in $S$, we sample a random value $v$ for it. The resulting set of values $V$, will determine uniquely the binary task.

Before selecting $v \in V$ we quantize the possible choices corresponding to factors of variations which may have more than six values to 2. We remark that this quantization affect only the task label definition.
For examples for x axis factor, we consider the object to be on the left if its x coordinate is less than the medial axis of the image, on the right otherwise. The \texttt{DSprites} dataset has the following set of factors of variations $Z_{dsprites}=\{shape,size, angle,x_{pos},y_{pos}\}$ and example of task is \textit{There is a big object on the right} where $k_t=2$ the affected factors are   $size, x_{pos}$. Another example is \textit{There is a small heart on the top left }, where $k_t=4$ the affected factors are   $shape,size, x_{pos},y_{pos}$. Obervations are labelled positively of negatively if their corresponding factors of variations matching in the values with the one specified by the current task.

We then samples random query $Q$ and support $U$ set  of samples balanced with respect to postive and negative labels of task  task $t$, using stratified sampling.

\subsection{Experiments on domain shifts}
For the domain generalization and few-shot transfer learning experiments we put ourselves in the same settings of \cite{gulrajani2020search,koh2021wilds} to ensure a fair comparison. Namely, for each dataset we use the same augmentations, and same backbone models. 

For solving the inner problem in Equation \ref{eq:inner_problem}, we used Adam optimizer \cite{kingma2014adam}, with a learning rate of $1e-2$, momentum $0.99$, with the number of gradient steps varying from $50$ to $100$, in domain shifts experiments.

\textbf{Task generation}. The task (or episode) sampling procedure is done as follows: each task is a multiclass classification problem: we set the number of classes $C$ to $C=5$ when the original number of classes $K_{train}$ in the dataset is higher than five, i.e. $K_{train}>5$. Otherwise we set $C=K_{train}$. During training, the sizes of the support set $U$ and query sets $Q$ where set to $|U|=25, |Q|=15$ similar to as done in prior meta-learning literature \cite{lee2019meta, dhillon2019baseline}. Changing these parameters has similar effects from what has been observed in many meta learning approaches(e.g. \cite{lee2019meta, dhillon2019baseline}). 

For binary datasets such as Camelyon17 or Waterbirds the possible classes to be predicted are always the same across tasks: what is changing is the composition of $U$ and $Q$. Keeping their cardinality low, we ensure that some tasks will not contain spurious correlation that may be present in the dataset, while other ones will still retain it, and the regularizers will satisfy solutions which discards the spurious information. We can observe evidence of this in the experimental results in Tables \ref{tab:camelyon}, \ref{tab:dg_domainbed} and qualitatively in Figure \ref{fig:waterbirds_qualitative}.

\subsection{Selection of $\alpha$ and $\beta$}
To find the best regularization parameters $\alpha,\beta$ weighting the sparsity and feature sharing regularizers in Equation~\ref{eq:regularizer} respectively, we perform model selection according to the highest accuracy on a validation set. We report in Table \ref{tab:model_selection} the value selected for each experiment.
\begin{table}[ht!]
    \caption{Selected values for $\alpha$ and $\beta$ for all experiments, applying model selection on validation set.}
    \centering
    \begin{tabular}{ccc}
\toprule 
\textbf{Experiment} & $\mathbf{\alpha}$ & $\mathbf{\beta}$  \\
\midrule 
Table 1 	& 1e-2 & 	0.15 \\
Table 2 	& 1e-2 	&5e-2 \\
Table 3 	& 2.5e-3 &	5e-2 \\
Table 4 	& 1.5e-3 &	1e-2 \\
Table 5,  6 	& 2.5e-3 &	1e-2 \\
Table 7 	& 2.5e-3 &	1e-2 \\
\bottomrule
    \end{tabular}

    \label{tab:model_selection}
\end{table}
\section{Additional results} \label{sec:app_additional_results}

\subsection{Synthetic experiments}

\textbf{Enforcing disentanglement}: In Table \ref{tab:full_results_synthetic} we report diverse disentanglement scores (DCI disentanglement, DCI completeness, DCI informativeness) on  the \texttt{DSprites}, \texttt{3DShapes}, \texttt{SmallNorb},\texttt{Cars} datasets, showing that the sparsity and feature sharing regularizers effectively enforce disentanglement.

\begin{table}[!ht]
    \centering
    \caption{\textit{Enforcing disentanglement}. DCI \cite{eastwood2018} disentanglement, completeness and informativeness scores and ID accuracy on test samples for a model trained without enforcing sufficiency and minimality (top row), and model with the regularizers activated (bottom row). While attaining similar performance on accuracy, the model with the activated regularizer always show higher disentanglement. See Table for additional scores.}
    \begin{tabular}{ccccc}
    \toprule 
         & DSprites & 3DShapes& SmallNorb & Cars \\ 
        \midrule 
        \textit{Without regularization} &   &   &   &   \\ 
        DCI Disentanglement & 16.6 & 44.4 & 16.5 & 60.5 \\ 
        DCI Completeness & 17.5 & 39.1 & 12.9 & 50.8 \\ 
        DCI Informativeness & 88.0 & 87.6 & 90.5 & 95.5 \\ 
                  \midrule 
        \textit{With regularization} &   &   &   &   \\ 
        DCI Disentanglement & 69.9 & 87.7 & 60.5 & 92.3 \\ 
        DCI Completeness & 72.3 & 88.4 & 63.2 & 57.1 \\ 
        DCI Informativeness & 96.0 & 95.7 & 95.4 & 99.7 \\ 
        \bottomrule
    \end{tabular}
    \label{tab:full_results_synthetic}
\end{table}

\begin{figure}
\centering
\begin{overpic}[width=0.6\linewidth,trim={0.45cm 0 1.1cm 1.1cm},clip]{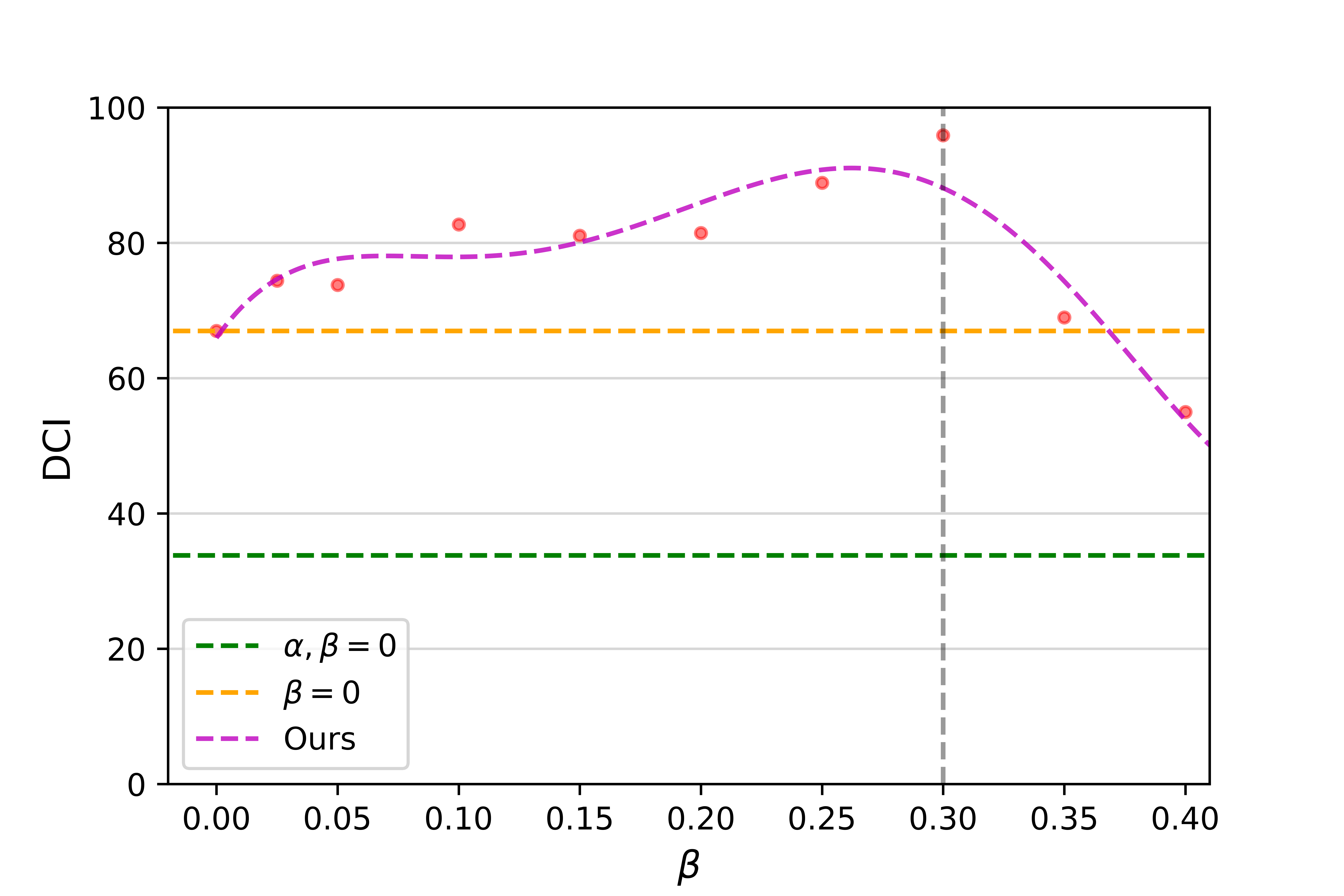}
    \put(26.3,16.){ \tiny{\cite{lachapelle2022synergies}}}
    %\put(20,20){\color{green}{-} \color{black} \footnotesize $\alpha=0,\beta=0$}
    \end{overpic}
    \captionof{figure}{\looseness=-1\textit{Role of minimality (3DShapes)}: We plot the DCI disentanglement metric of a set of models (\textit{red dots}) trained on fixed tasks from \texttt{3Dshapes}: Training without regularizers leads to no disentanglement (\textit{green}). Enforcing sparsity alone (\textit{yellow}, akin to \cite{lachapelle2022synergies}) achieves good disentanglement ($DCI=67.0$), but some features may be split or duplicated. Enforcing both minimality and sparse sufficiency (\textit{magenta}) attains the best $DCI$ ($95.9$). When $\beta$ is too high ($>0.25$) activated features collapses into few clusters with respect to tasks.} %For exact values and qualitative evidence see Table \ref{tab:betavsdci} and Figure \ref{fig:disentanglement_qualitative} in Appendix.}
 \label{fig:betavsDCI3DShapes}

 \end{figure}
\begin{figure}[h!]
\centering
\begin{overpic}[width=0.7\linewidth,trim={10cm 2.8cm 9cm 2cm},clip]{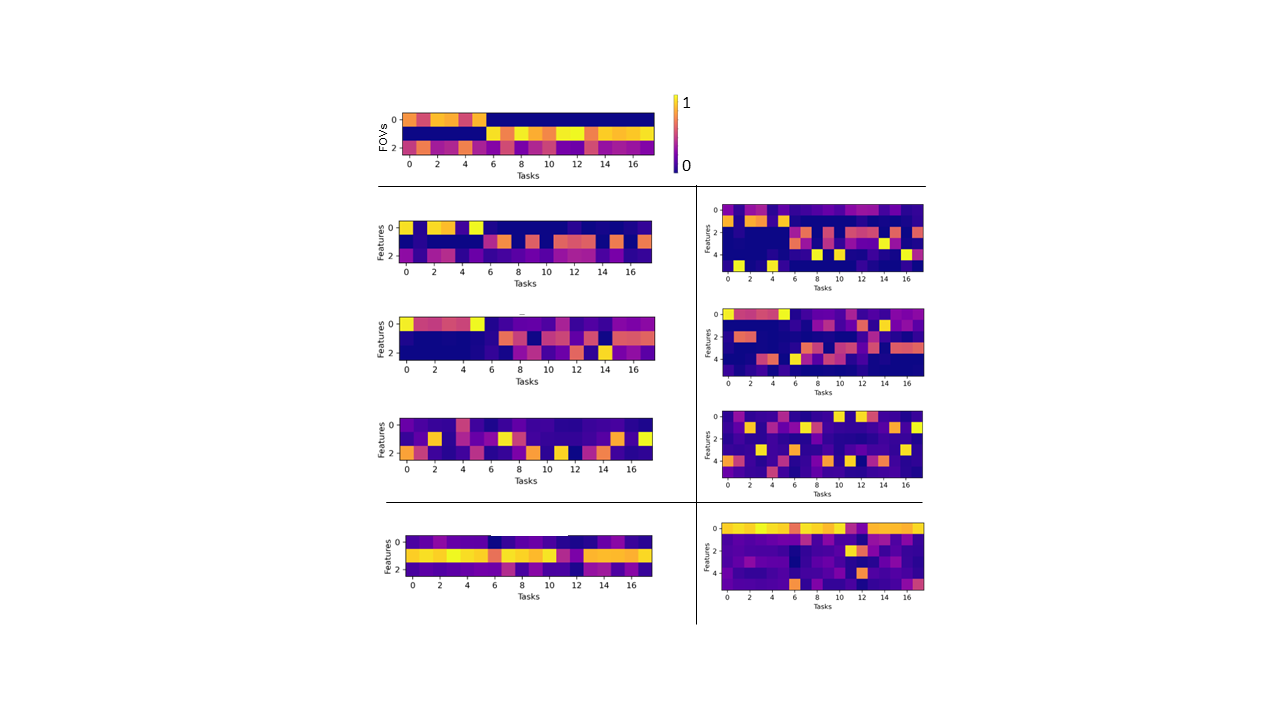}
\put(15,15) {\tiny$\beta=0.4, DCI=30.5$ }
\put(10,36) {\tiny$\beta=0, \alpha=0, DCI=27.8$ }
\put(15,54) {\tiny$\beta=0, DCI=71.9$ }
\put(15,71) {\tiny$\beta=0.2, DCI=98.8$ }
\end{overpic}
\caption{Qualitative dependency of disentanglement from the weight of our penalties ($\alpha=0.01$ unless otherwise specified). The model that attains the best disentanglement ($DCI=98.8$) uses both. \textit{ Left column, top}: ground-truth importance weights of each latent factor for each task. \textit{Right column}: we train models with different $\beta$ and visualize the weights assigned to each learned feature on each task. \textit{Left column}: to determine whether the model recover the ground-truth latents, we select the 3 top features and compare their assigned weights on different tasks with the ground-truth weights. \textit{Bottom row}: example of a failure case with high $\beta$.} \label{fig:disentanglement_qualitative}

\end{figure}

\textbf{The role of minimality}.
In Figure \ref{fig:disentanglement_qualitative} we show the qualitative results accompanying Figure \ref{fig:betavsDCI}.
The qualitative results in the Figure are produced visualizing  matrices of feature importance~\cite{locatello2020sober} computed fitting Gradient Boosted Trees (GBT) on the learned representations w.r.t. task labels, and on the factors of variations w.r.t. task labels and compare the results. In each matrix the x axis represents the tasks and the y axis the features, and each entries the amount of feature importance (which goes from 0 to 1).
In Figure \ref{fig:betavsDCI3DShapes} we show the same experiment on the  \texttt{3DShapes} dataset.

\textbf{Task compositional generalization}.
In Table \ref{tab:task_generalization} we show the quantitative results accompanying Figure \ref{fig:task_composition}.

\begin{table}%[ht!]
    \caption{Quantitative results accompanying Figure \ref{fig:disentanglement_qualitative}}
    
    \centering
    \resizebox{0.8\linewidth}{!}{
    \begin{tabular}{ccccc}
    \toprule 
        & $\alpha=0, \beta=0$ & $\alpha=1e-2, \beta=0$ & $\alpha=1e-2, \beta=0.2$ & $\alpha=1e-2, \beta=0.4$\\
    \midrule 
           DCI&   27.8       &  71.9         & 98.8  & 30.5 \\
    \bottomrule 
    \end{tabular}}

    \label{tab:betavsdci}
\end{table}
\begin{table}[h!]
    \centering
    \caption{\emph{Task compositional generalization}: Mean accuracy over 100 random tasks reported for group of tasks of growing support (\textit{second, third, fourth column}) for  a model trained without inductive biases (\textit{top row}) and enforcing them (\textit{bottom row}). The latter show better compositional generalization resulting from the properties enforced on the representation}
    \label{tab:task_generalization}
    \begin{tabular}{cccccc}
    \toprule 
      & Acc ID & DCI  & $|S|=3 $& $|S|=4 $& $|S|=5$ \\
    \midrule
      \emph{No reg} & 88.7 & 22.8 &  72.6      & 63.3    &  59.9         \\
    \midrule 
     $\alpha,\beta$  & $\mathbf{93.2}$ & $\mathbf{59.4}$ & $\mathbf{83.0}$      & $\mathbf{78.8}$   &    $\mathbf{76.8}$       \\
    \bottomrule
    \end{tabular}
\end{table}

\subsection{Properties of the learned representations} \label{sec:app_properties_analysis}
\begin{figure}[h]
\centering
\resizebox{0.9\linewidth}{!}{
\begin{tabular}{l|l}
%  \includegraphics[width=0.5\linewidth]{pictures/a.png}
% & \includegraphics[width=0.5\linewidth]{pictures/b.png} \\
%  \includegraphics[width=0.5\linewidth]{pictures/bird.png}
% & \includegraphics[width=0.5\linewidth]{pictures/bird_erm.png} \\
%  \includegraphics[width=0.5\linewidth]{pictures/bird2.png}
% & \includegraphics[width=0.5\linewidth]{pictures/bird2erm.png} \\
 \includegraphics[width=0.45\linewidth]{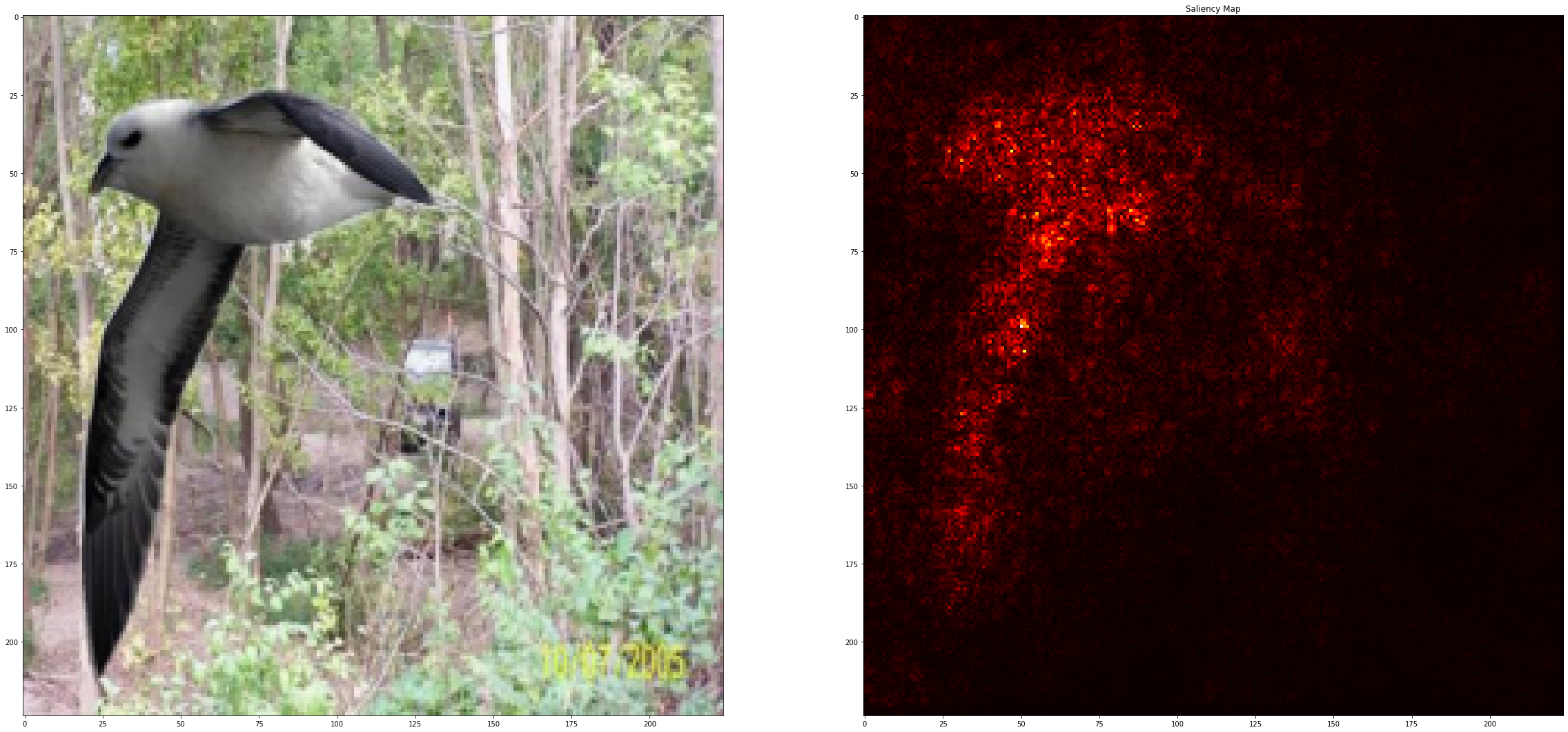}
& \includegraphics[width=0.45\linewidth]{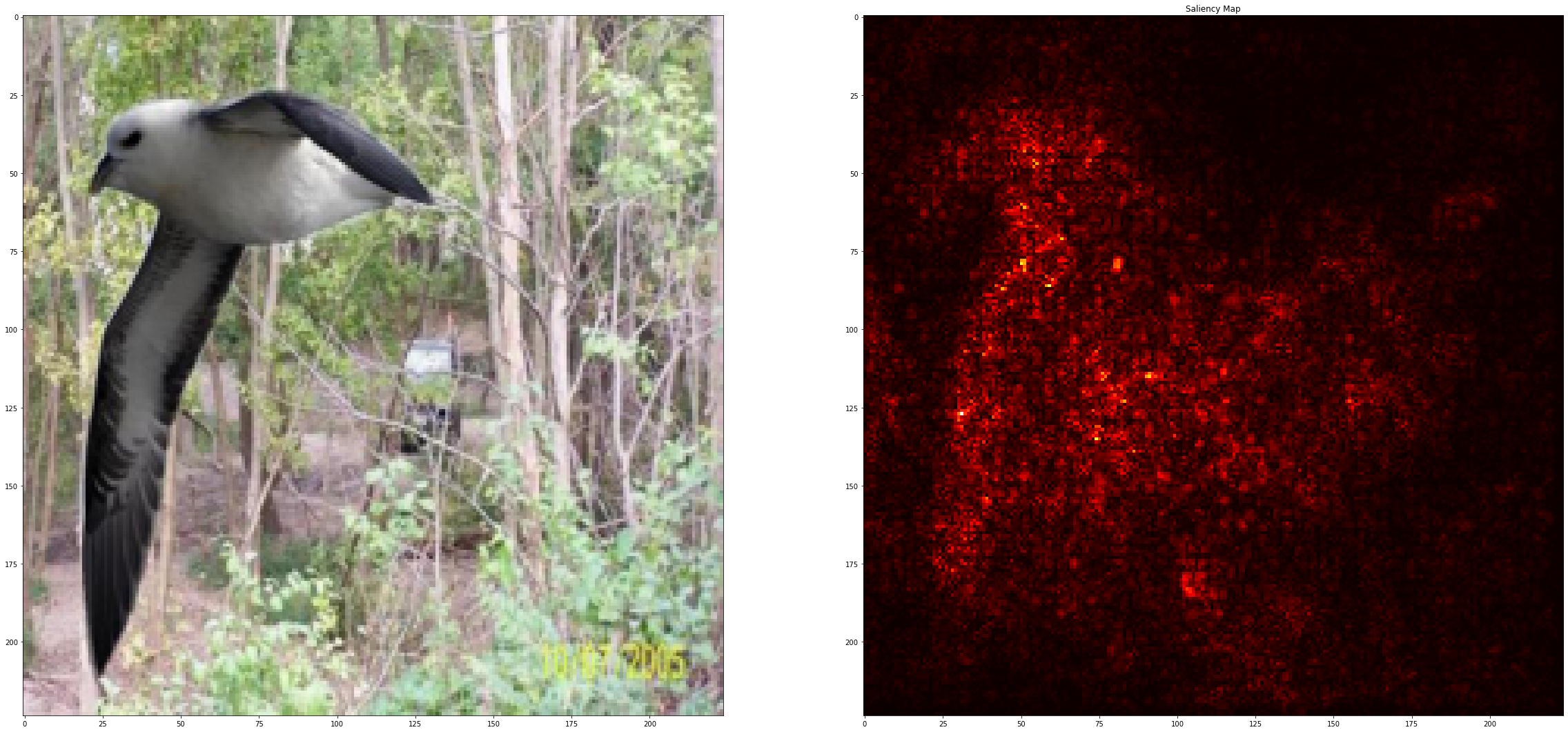} \\
\hline
 \includegraphics[width=0.45\linewidth]{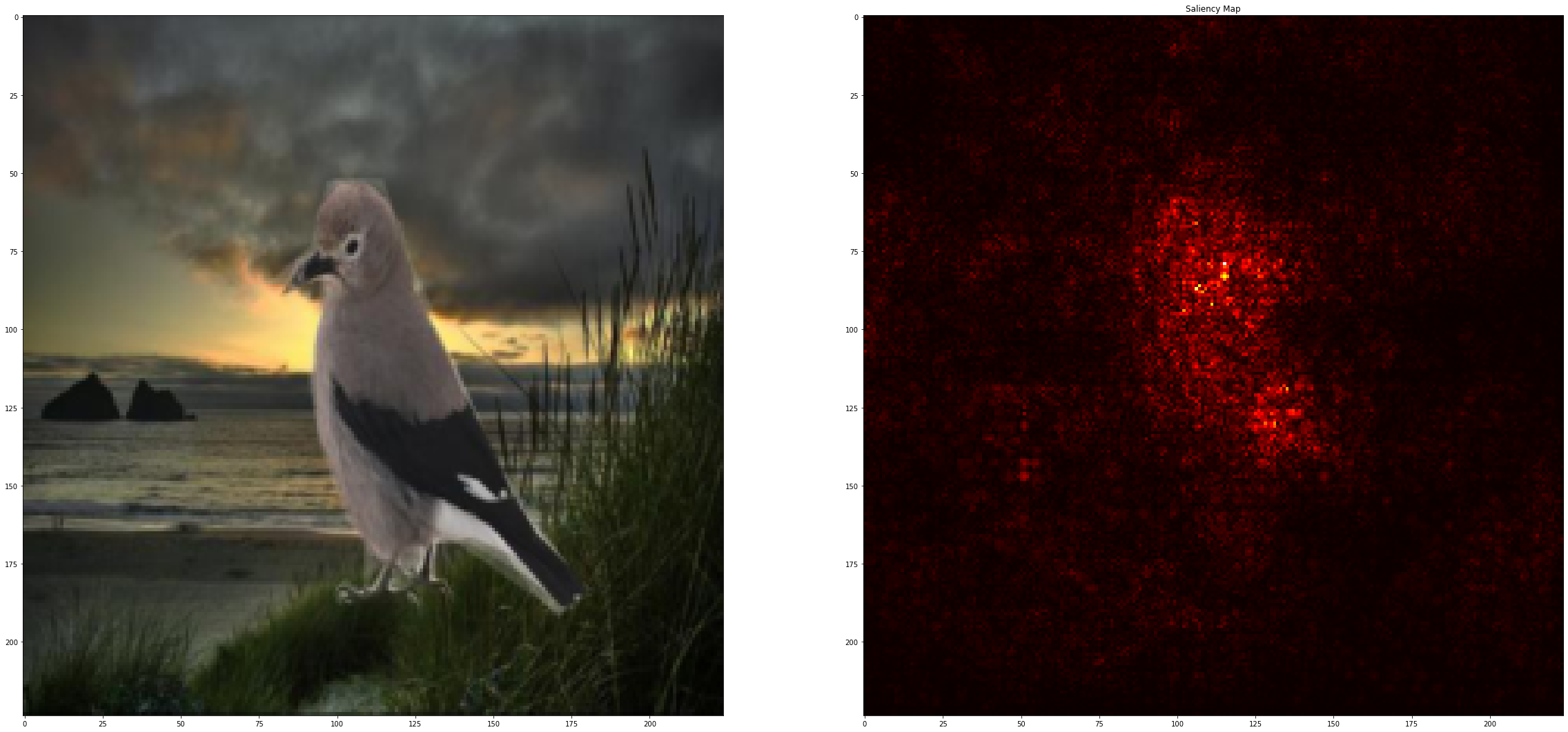}
& \includegraphics[width=0.45\linewidth]{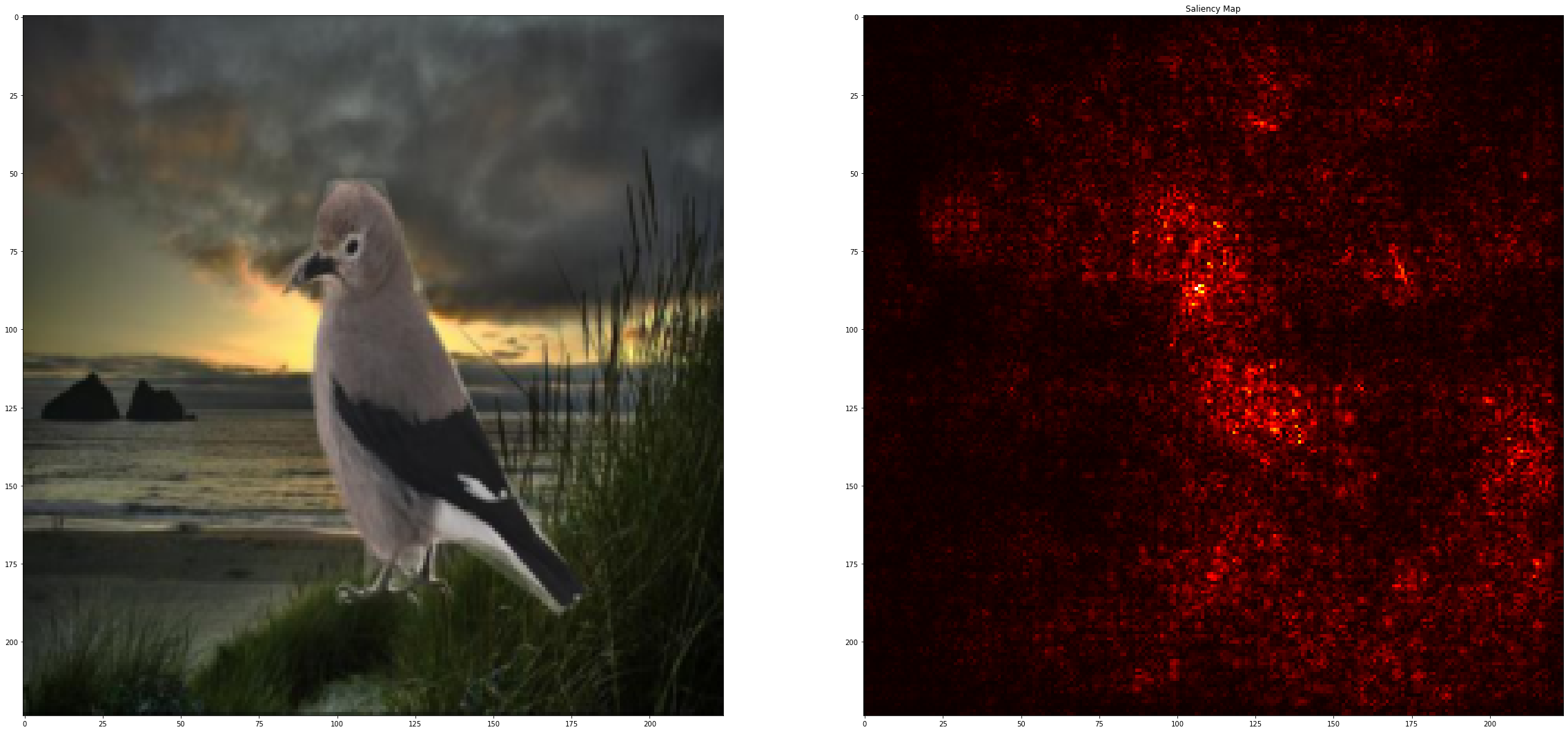} \\
\end{tabular}}
\caption{\emph{Feature sufficiency:} \textit{Left}, pairs of random samples and saliency maps computed on activations with our method. All samples are correctly classified. \textit{Right}, corresponding saliency maps \cite{adebayo2018sanity} an ERM based method: the first row is misclassifed by the network, the last  is correctly classified. The ERM model depends on features from the background, resulting in a higher prediction error on mixed subdomains. Our model is robust to spurious correlations and satisfies the sufficiency assumptions.}
\label{fig:waterbirds_qualitative}
\end{figure}

%\FL{I find this whole section confusing. Maybe move to the appendix entirely? We could turn this into an ablation / deep dive, where we also add the beta greater than 0 at test time.}
\textbf{Feature sufficiency.}
The sufficiency property is crucial for robustness to spurious correlations in the data. If the model can learn and select the relevant features for a task, while ignoring the spurious ones, sufficiency is satisfied, resulting in robust performance under subpopulation shifts, as shown in Tables \ref{tab:civilcomments} and \ref{tab:dg_domainbed}.
To get qualitative evidence of the sufficiency in the representations, in Figure \ref{fig:waterbirds_qualitative} we show the saliency maps computed from the activations of our model and a corresponding model trained with ERM. Our model can learn features specific to the subject of the image, which are relevant for classification, while ignoring background information. This can be observed in both correctly classified (bottom row) and misclassified (top row) samples by ERM. In contrast, ERM activates features in the background and relies on them for prediction.

\begin{figure}[ht!]
    \centering 
    \includegraphics[width=\linewidth]{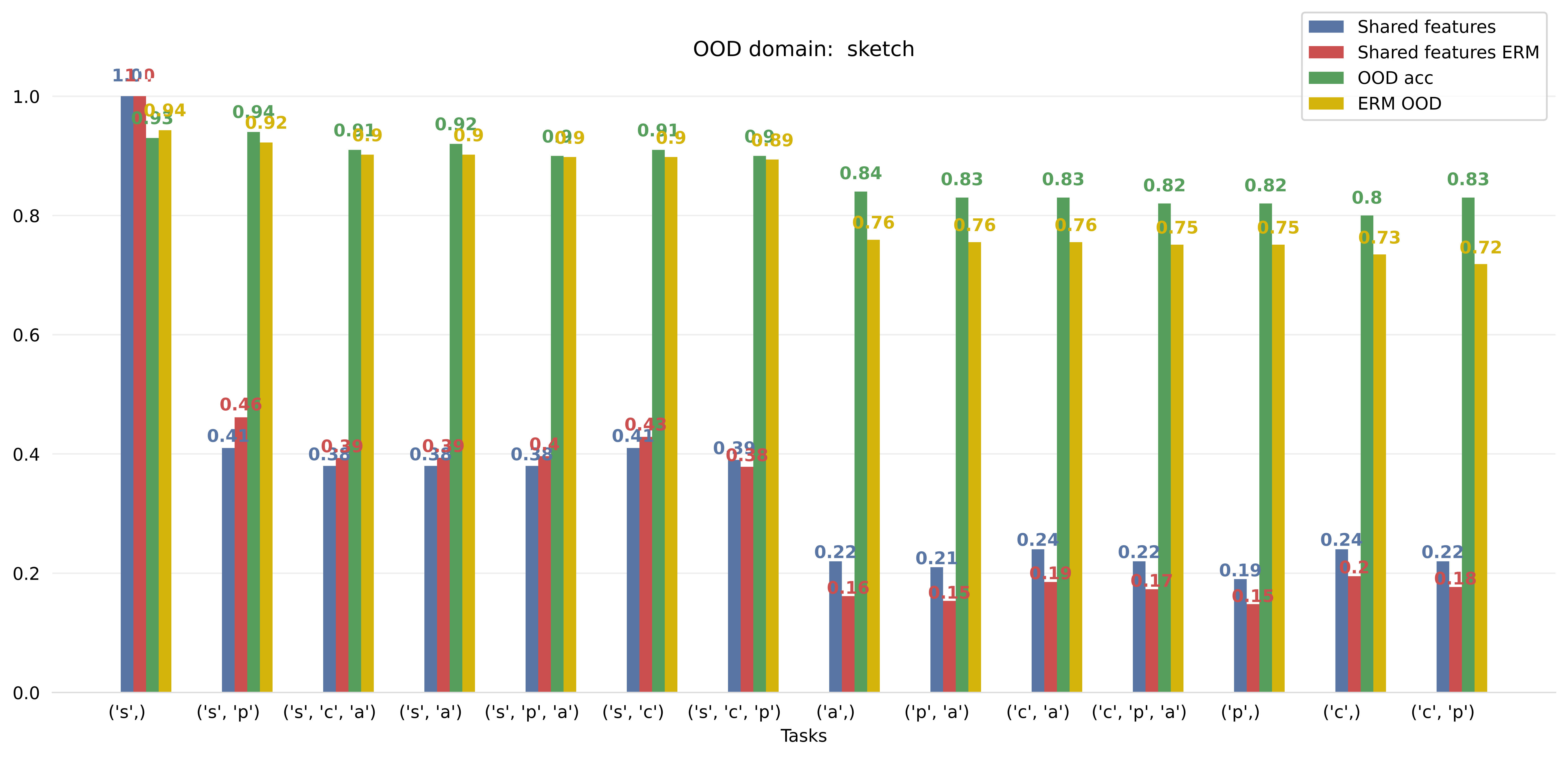}
    \caption{\emph{Fraction of shared features VS accuracy}. Barplot of OOD accuracies on the \textit{Sketch} domain for our model (green) and ERM (yellow) on the 14 tasks sampled from \texttt{PACS}, along with the fraction of shared features with the OOD domain for each task (blue for our model, red for ERM). Each task is sampled from a single domain  or from the intersections of domains. Tasks are labelled according to the sampling domain on the x axis. The fraction of shared features and OOD accuracy have a correlation coefficient of $97.5$.}
    \label{fig:fract_shared_sketch}
\end{figure}

\begin{figure}[ht!]
    \centering
    \includegraphics[width=\linewidth,trim={0 0.2cm 0 0},clip]{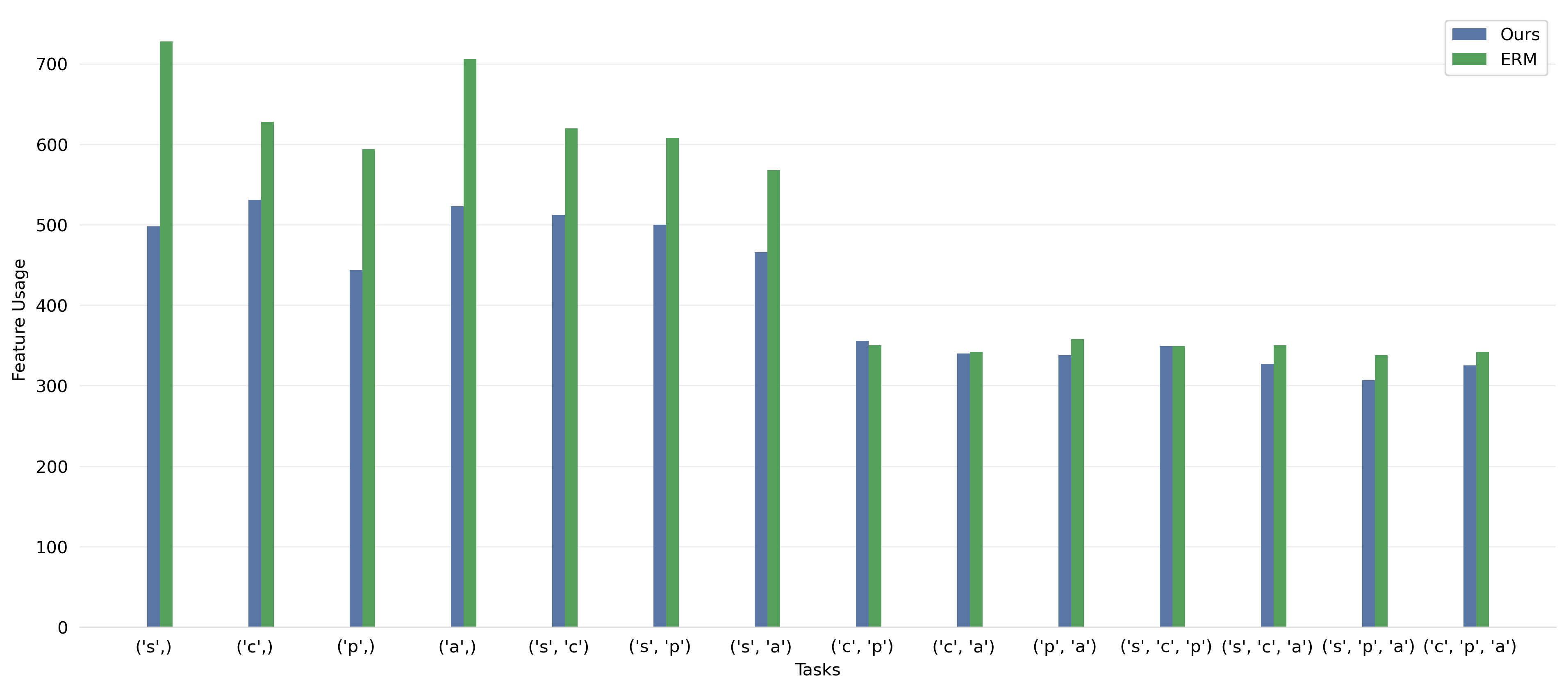}
    
    \caption{Barplot of feature usage (number of activated features) for each task for our model (blue) and ERM model (green) referring to the experiment in Figure \ref{fig:fract_shared_sketch}. Our method uses fewer features than ERM while also generalizing better.}
    \label{fig:feature_usage}
\end{figure}

\textbf{Feature sharing.}
In this section, we study the minimality properties of the representations learned by our method. To achieve this, we conduct the following experiment. We randomly draw 14 tasks from the $\sum_{i=1}^{3}\binom{4}{i}$ possible combinations of the four domains in the \texttt{PACS} dataset. We use the data from these tasks to fit the linear head and test the model accuracy on the OOD domain (e.g. the \textit{sketch} domain).
In Figure \ref{fig:fract_shared_sketch}, we show the performance on each task, ordered on the x axis according to OOD accuracy of a model trained with ERM (in yellow). We also report the fraction of activated features (in blue) shared between each task and the OOD task, and the same(red) for the ERM model. The fraction of activated features is computed by looking at the matrix of coefficients of the sparse linear head $\phi \in \mathbb{R}^{M \times C}$, where $M$ is the number of features and $C$ the number of classes, after fitting on each task. Specifically, is computed as $\frac{\sum_m \left[\tilde{\phi}_\epsilon \cap \tilde{\phi}_\epsilon^{OOD}  \right]}{\sum_m \left[\tilde{\phi}_\epsilon \cup \tilde{\phi}_\epsilon^{OOD}\right]}$ where $\tilde{\phi}_\epsilon=\frac{1}{C}\sum_c |\phi_{m,c}|> \epsilon$ and $\phi^{OOD}$ is the matrix of coefficient of the OOD task. We set $\epsilon=0.01$. 
From Figures~\ref{fig:fract_shared_sketch} and~\ref{fig:feature_usage} we draw the following conclusions: (i) When the accuracy of the ERM decreases (i.e., the current task is farther from the OOD test task), our method is still able to retain a high and consistent accuracy, demonstrating that our features are more robust out-of-distribution. This is further supported by the higher number of shared features compared to ERM, as we move away from the testing domain. (ii) The correlation between the fraction of shared features and the accuracy OOD demonstrates that the method is able to learn general features that transfer well to unseen domains, thanks to the minimality constraint. Additionally, this measure serves as a reliable indicator of task distance, as discussed in the next section. (iii) Even though the same sparse linear head is used on top of the ERM and our features, our method is able to achieve better OOD performance with fewer features, further demonstrating our feature minimality.

% \begin{table}[h!]
%     \centering
%     \begin{tabular}{c|c}
%         \includegraphics[width=0.45\linewidth]{pictures/oty0iy1m0.01sketchnex_150_mean_before__accVSshared.png}      & \includegraphics[width=0.45\linewidth]{pictures/oty0iy1m0.01sketchnex_150_mean_before__featureusage.png}
%     \end{tabular}
%     \caption{Caption feat sharing}
%     \label{fig:feat_sharing}
% \end{table}

\subsection{CivilComments}
See Table \ref{tab:civilcomments} for the quantitative results accompanying to Figure \ref{fig:civilcomments} in the paper and \ref{tab:civilcommentsgroups} for result on groups on the civil comments dataset.

\begin{table}[ht!]
    \caption{\textit{Quantitative results on CivilComments}: we report the accuracy on test averaged across all demographic groups (\textit{left}), and the worst group accuracy (\textit{right}). We show that our method performs similarly in terms of average accuracy and outperforms in terms of worst group accuracy, without using any knowledge on the group composition in the training data. This Table accompanies Figure \ref{fig:civilcomments}}
    \label{tab:civilcomments}
    \centering
    \begin{tabular}{ccc}
    \toprule 
      & avg acc  & worst group  acc\\
    \midrule 
      ERM  & $\mathbf{92.2}$ & 56.5   \\
      DRO  &  90.2 & 69  \\
      Ours &  91.2 \footnotesize$\pm$ 0.2 & $\mathbf{75.45}$\footnotesize$\pm$  0.1\\
        \bottomrule 
    \end{tabular}
\end{table}

\begin{table}[ht!]
    \caption{Civilcomments quantitative results pergroup.}
    \centering
    \resizebox{\linewidth}{!}{
    \begin{tabular}{ccccccccc}
    \hline 
         & Male & Female & LGBTQ & Christian & Muslim & Other religion & Black & White \\
    \hline 
    \emph{GroupDRO}& & & & & & & & \\
     Toxic& $ 75.1\footnotesize \pm2.1 $  & $73.7 \footnotesize \pm 1.5$ & $73.7  \footnotesize \pm 4 $ & $ 69.2\footnotesize \pm 2.0$ & $ 72.1\footnotesize \pm 2.6 $ & $ 72.0 \footnotesize \pm 2.5 $ & $79.6 \footnotesize \pm 2.2 $ & $ 78.8 \footnotesize \pm 1.7 $     \\
     Non Toxic& $ 88.4\footnotesize \pm0.7 $  & $90.0 \footnotesize \pm 0.6$ & $76.0 \footnotesize \pm 3.6 $ & $ 92.6\footnotesize \pm 0.6$ & $ 80.7\footnotesize \pm 1.9 $ & $ 87.4 \footnotesize \pm 0.9 $ & $72.2 \footnotesize \pm 2.3 $ & $ 73.4 \footnotesize \pm 1.4 $     \\

    \emph{Ours}& & & & & & & & \\
     Toxic &$87.94 \footnotesize \pm 0.07$  & $ 89.17\footnotesize \pm0.05 $ & $ 77.25\footnotesize \pm 0.16$ & $92.25 \footnotesize \pm 0.16 $ & $ 80.6\footnotesize \pm0.29 $ & $87.79\footnotesize \pm 0.26 $ & $ 75.45\footnotesize \pm 0.17 $ & $ 78.35\footnotesize \pm 0.02$     \\
     Non toxic &$91.62\footnotesize \pm 0.11$  & $ 91.52\footnotesize \pm0.11 $ & $ 91.71\footnotesize \pm 0.16$ & $91.11 \footnotesize \pm 0.1 $ &$91.81 \footnotesize \pm 0.12 $ & $ 91.32\footnotesize \pm0.1 $ &  $ 90.82\footnotesize \pm 0.12 $ & $ 92.04\footnotesize \pm 0.11$     \\
    \hline 
    \end{tabular}}    \label{tab:civilcommentsgroups}
\end{table}

\subsection{Full results Domain generalization} \label{sec:app_full_Dg_exp}
We report here comparison with several methods in the domain generalization literature, namely \cite{yan2020improve,blanchard2021domain,li2018learning,li2018domain,ganin2016domain,li2018deep,nam2021reducing,zhang2021adaptive,huang2020self,krueger2021out}.
\subsubsection{VLCS}

\begin{center}

\begin{tabular}{lccccc}
\toprule
\textbf{Algorithm}   & \textbf{C}           & \textbf{L}           & \textbf{S}           & \textbf{V}           & \textbf{Avg}         \\
\midrule
ERM                  & 97.7 $\pm$ 0.4       & 64.3 $\pm$ 0.9       & 73.4 $\pm$ 0.5       & 74.6 $\pm$ 1.3       & 77.5                 \\
IRM                  & 98.6 $\pm$ 0.1       & 64.9 $\pm$ 0.9       & 73.4 $\pm$ 0.6       & 77.3 $\pm$ 0.9       & 78.5                 \\
GroupDRO             & 97.3 $\pm$ 0.3       & 63.4 $\pm$ 0.9       & 69.5 $\pm$ 0.8       & 76.7 $\pm$ 0.7       & 76.7                 \\
Mixup                & 98.3 $\pm$ 0.6       & 64.8 $\pm$ 1.0       & 72.1 $\pm$ 0.5       & 74.3 $\pm$ 0.8       & 77.4                 \\
MLDG                 & 97.4 $\pm$ 0.2       & 65.2 $\pm$ 0.7       & 71.0 $\pm$ 1.4       & 75.3 $\pm$ 1.0       & 77.2                 \\
CORAL                & 98.3 $\pm$ 0.1       & \textbf{66.1} $\pm$ 1.2       & 73.4 $\pm$ 0.3       & 77.5 $\pm$ 1.2       & \textbf{78.8}                 \\
MMD                  & 97.7 $\pm$ 0.1       & 64.0 $\pm$ 1.1       & 72.8 $\pm$ 0.2       & 75.3 $\pm$ 3.3       & 77.5                 \\
DANN                 & 99.0 $\pm$ 0.3       & 65.1 $\pm$ 1.4       & 73.1 $\pm$ 0.3       & 77.2 $\pm$ 0.6       & 78.6                 \\
CDANN                & 97.1 $\pm$ 0.3       & 65.1 $\pm$ 1.2       & 70.7 $\pm$ 0.8       & 77.1 $\pm$ 1.5       & 77.5                 \\
MTL                  & 97.8 $\pm$ 0.4       & 64.3 $\pm$ 0.3       & 71.5 $\pm$ 0.7       & 75.3 $\pm$ 1.7       & 77.2                 \\
SagNet               & 97.9 $\pm$ 0.4       & 64.5 $\pm$ 0.5       & 71.4 $\pm$ 1.3       & 77.5 $\pm$ 0.5       & 77.8                 \\
ARM                  & 98.7 $\pm$ 0.2       & 63.6 $\pm$ 0.7       & 71.3 $\pm$ 1.2       & 76.7 $\pm$ 0.6       & 77.6                 \\
VREx                 & 98.4 $\pm$ 0.3       & 64.4 $\pm$ 1.4       & 74.1 $\pm$ 0.4       & 76.2 $\pm$ 1.3       & 78.3                 \\
RSC                  & 97.9 $\pm$ 0.1       & 62.5 $\pm$ 0.7       & 72.3 $\pm$ 1.2       & 75.6 $\pm$ 0.8       & 77.1                 \\
\textbf{Ours}       &  \textbf{98.1}$\pm$ 0.2 &      63.4$\pm$ 0.5 & \textbf{73.9} $\pm$ 0.8  & \textbf{78.2} $\pm$  0.7 & 78.4 \\
\bottomrule
\end{tabular}
\end{center}

\subsubsection{PACS}

\begin{center}
\begin{tabular}{lccccc}
\toprule
\textbf{Algorithm}   & \textbf{A}           & \textbf{C}           & \textbf{P}           & \textbf{S}           & \textbf{Avg}         \\
\midrule
ERM                  & 84.7 $\pm$ 0.4       & 80.8 $\pm$ 0.6       & 97.2 $\pm$ 0.3       & 79.3 $\pm$ 1.0       & 85.5                 \\
IRM                  & 84.8 $\pm$ 1.3       & 76.4 $\pm$ 1.1       & 96.7 $\pm$ 0.6       & 76.1 $\pm$ 1.0       & 83.5                 \\
GroupDRO             & 83.5 $\pm$ 0.9       & 79.1 $\pm$ 0.6       & 96.7 $\pm$ 0.3       & 78.3 $\pm$ 2.0       & 84.4                 \\
Mixup                & 86.1 $\pm$ 0.5       & 78.9 $\pm$ 0.8       & 97.6 $\pm$ 0.1       & 75.8 $\pm$ 1.8       & 84.6                 \\
MLDG                 & 85.5 $\pm$ 1.4       & 80.1 $\pm$ 1.7       & 97.4 $\pm$ 0.3       & 76.6 $\pm$ 1.1       & 84.9                 \\
CORAL                & 88.3 $\pm$ 0.2       & 80.0 $\pm$ 0.5       & 97.5 $\pm$ 0.3       & 78.8 $\pm$ 1.3       & 86.2                 \\
MMD                  & 86.1 $\pm$ 1.4       & 79.4 $\pm$ 0.9       & 96.6 $\pm$ 0.2       & 76.5 $\pm$ 0.5       & 84.6                 \\
DANN                 & 86.4 $\pm$ 0.8       & 77.4 $\pm$ 0.8       & 97.3 $\pm$ 0.4       & 73.5 $\pm$ 2.3       & 83.6                 \\
CDANN                & 84.6 $\pm$ 1.8       & 75.5 $\pm$ 0.9       & 96.8 $\pm$ 0.3       & 73.5 $\pm$ 0.6       & 82.6                 \\
MTL                  & 87.5 $\pm$ 0.8       & 77.1 $\pm$ 0.5       & 96.4 $\pm$ 0.8       & 77.3 $\pm$ 1.8       & 84.6                 \\
SagNet               & 87.4 $\pm$ 1.0       & 80.7 $\pm$ 0.6       & 97.1 $\pm$ 0.1       & 80.0 $\pm$ 0.4       & 86.3                 \\
ARM                  & 86.8 $\pm$ 0.6       & 76.8 $\pm$ 0.5       & 97.4 $\pm$ 0.3       & 79.3 $\pm$ 1.2       & 85.1                 \\
VREx                 & 86.0 $\pm$ 1.6       & 79.1 $\pm$ 0.6       & 96.9 $\pm$ 0.5       & 77.7 $\pm$ 1.7       & 84.9                 \\
RSC                  & 85.4 $\pm$ 0.8       & 79.7 $\pm$ 1.8       & 97.6 $\pm$ 0.3       & 78.2 $\pm$ 1.2       & 85.2                 \\
\textbf{Ours}       & \textbf{86.7}  $\pm$ 0.1&  \textbf{83.5}     $\pm$ 0.8 &  \textbf{97.8} $\pm$ 0.1  & \textbf{83.1} $\pm$ 0.1 & \textbf{87.5} \\

\bottomrule
\end{tabular}
\end{center}

\subsubsection{OfficeHome}

\begin{center}
\begin{tabular}{lccccc}
\toprule
\textbf{Algorithm}   & \textbf{A}           & \textbf{C}           & \textbf{P}           & \textbf{R}           & \textbf{Avg}         \\
\midrule
ERM                  & 61.3 $\pm$ 0.7       & 52.4 $\pm$ 0.3       & 75.8 $\pm$ 0.1       & 76.6 $\pm$ 0.3       & 66.5                 \\
IRM                  & 58.9 $\pm$ 2.3       & 52.2 $\pm$ 1.6       & 72.1 $\pm$ 2.9       & 74.0 $\pm$ 2.5       & 64.3                 \\
GroupDRO             & 60.4 $\pm$ 0.7       & 52.7 $\pm$ 1.0       & 75.0 $\pm$ 0.7       & 76.0 $\pm$ 0.7       & 66.0                 \\
Mixup                & 62.4 $\pm$ 0.8       & 54.8 $\pm$ 0.6       & 76.9 $\pm$ 0.3       & 78.3 $\pm$ 0.2       & 68.1                 \\
MLDG                 & 61.5 $\pm$ 0.9       & 53.2 $\pm$ 0.6       & 75.0 $\pm$ 1.2       & 77.5 $\pm$ 0.4       & 66.8                 \\
CORAL                & 65.3 $\pm$ 0.4       & 54.4 $\pm$ 0.5       & 76.5 $\pm$ 0.1       & 78.4 $\pm$ 0.5       & 68.7                 \\
MMD                  & 60.4 $\pm$ 0.2       & 53.3 $\pm$ 0.3       & 74.3 $\pm$ 0.1       & 77.4 $\pm$ 0.6       & 66.3                 \\
DANN                 & 59.9 $\pm$ 1.3       & 53.0 $\pm$ 0.3       & 73.6 $\pm$ 0.7       & 76.9 $\pm$ 0.5       & 65.9                 \\
CDANN                & 61.5 $\pm$ 1.4       & 50.4 $\pm$ 2.4       & 74.4 $\pm$ 0.9       & 76.6 $\pm$ 0.8       & 65.8                 \\
MTL                  & 61.5 $\pm$ 0.7       & 52.4 $\pm$ 0.6       & 74.9 $\pm$ 0.4       & 76.8 $\pm$ 0.4       & 66.4                 \\
SagNet               & 63.4 $\pm$ 0.2       & 54.8 $\pm$ 0.4       & 75.8 $\pm$ 0.4       & 78.3 $\pm$ 0.3       & 68.1                 \\
ARM                  & 58.9 $\pm$ 0.8       & 51.0 $\pm$ 0.5       & 74.1 $\pm$ 0.1       & 75.2 $\pm$ 0.3       & 64.8                 \\
VREx                 & 60.7 $\pm$ 0.9       & 53.0 $\pm$ 0.9       & 75.3 $\pm$ 0.1       & 76.6 $\pm$ 0.5       & 66.4                 \\
RSC                  & 60.7 $\pm$ 1.4       & 51.4 $\pm$ 0.3       & 74.8 $\pm$ 1.1       & 75.1 $\pm$ 1.3       & 65.5                 \\
\textbf{Ours}       &  \textbf{66.7} $\pm$ 0.1 &     \textbf{56.3}  $\pm$ 0.7 & \textbf{79.2} $\pm$ 0.5 &  \textbf{81.3} $\pm$ 0.4 &  \textbf{70.9}\\

\bottomrule
\end{tabular}
\end{center}
\subsection{Few-shot transfer learning} \label{sec:app_fewshot}
Results on few-shot transfer learning on datasets \texttt{PACS},\texttt{VLCS},\texttt{OfficeHome},\texttt{Waterbirds} in  Tables \ref{tab:da_pacs},\ref{tab:da_vlcs},\ref{tab:da_offhome} and 
\ref{tab:da_waterbirds}.

\begin{table}[ht!]
\caption{Results few-shot transfer learning on PACS}
\centering
\begin{tabular}{cccccc}
\toprule
\multicolumn{2}{l}{\textbf{Dataset/Algorithm}} & \multicolumn{3}{c}{\textbf{OOD accuracy (by domain)}} \\
\midrule
{\textbf{PACS 1-shot}} &         S &           A &   P &        C & Average\\

ERM       &  72.3 \footnotesize$\pm$ 0.3 &      $80.4$ \footnotesize$\pm$  0.09 &  93.3 \footnotesize$\pm$ 4.1 & 75.8\footnotesize$\pm$ 2.6 & 80.5\\
Ours       &  $\mathbf{75.4}$ \footnotesize$\pm$ 3&      $\mathbf{81.7}$\footnotesize$\pm$ 0.8 &  $\mathbf{98.0}$ \footnotesize$\pm$ 0.8  &  $\mathbf{71}$ \footnotesize$\pm$ 5.2 & $\mathbf{81.5}$ \\
\midrule

{\textbf{PACS 5-shot}} &         S &           P &   A &        C & Average\\

ERM       &  84.9\footnotesize$\pm$ 1.1 &      85.7 \footnotesize$\pm$  0.08 &  98.6 \footnotesize$\pm$ 0.0 & 79.1 \footnotesize$\pm$ 0.9 & 87.1\\
Ours       &  $\mathbf{85.0}$ \footnotesize$\pm$ 0.1&      $\mathbf{86.7}$\footnotesize$\pm$ 0.8 &  $\mathbf{97.8}$ \footnotesize$\pm$ 0.1  &  $\mathbf{83.5}$ \footnotesize$\pm$ 0.1 & $\mathbf{88.3}$ \\
\midrule

{\textbf{PACS 10-shot}} &         S &           P &   A &        C & Average\\

ERM       &  81.0 \footnotesize$\pm$ 0.1 &      88.9 \footnotesize$\pm$  0.1 &  97.4 \footnotesize$\pm$ 0.0 & 84.2 \footnotesize$\pm$ 0.9 & 87.9\\
Ours       &  $\mathbf{86.2}$ \footnotesize$\pm$ 0.5&      $\mathbf{90.0}$ \footnotesize$\pm$ 0.8 &  $\mathbf{98.9}$ \footnotesize$\pm$ 0.1  &  $\mathbf{86.6}$ \footnotesize$\pm$ 0.1 & $\mathbf{90.4}$ \\
\bottomrule

\end{tabular}

\label{tab:da_pacs}
\end{table}

\begin{table}[h!]
\caption{results few-shot transfer learning on VLCS}
\centering
\begin{tabular}{cccccc}
\toprule
\multicolumn{2}{l}{\textbf{Dataset/Algorithm}} & \multicolumn{3}{c}{\textbf{OOD accuracy (by domain)}} \\
\midrule
{\textbf{VLCS 1-shot}} &                 C &           L &   V &        S & Average\\
ERM       &  98.9  \footnotesize$\pm$ 0.4&  32.7  \footnotesize$\pm$ 16.2&  59.8  \footnotesize$\pm$ 10.7&  47.5  \footnotesize$\pm$ 11.2 & 59.7 \\
Ours      & $\mathbf{ 98.6 }$ \footnotesize$\pm$ 0.3& $\mathbf{ 51.0 }$ \footnotesize$\pm$ 4.9& $\mathbf{ 61.2 }$ \footnotesize$\pm$ 9.8& $\mathbf{ 61.9 }$ \footnotesize$\pm$ 9.7 & $\mathbf{68.2}$ \\

\midrule

{\textbf{VLCS 5-shot}} &                  C &           L &   V &        S & Average\\
ERM       &  99.4  \footnotesize$\pm$ 0.2&  50.0  \footnotesize$\pm$ 6.2&  71.9  \footnotesize$\pm$ 3.2&  65.3  \footnotesize$\pm$ 2.8 & 71.7 \\
Ours       
& $\mathbf{ 98.9 }$ \footnotesize$\pm$ 0.4& $\mathbf{ 56.0 }$ \footnotesize$\pm$ 6.2& $\mathbf{ 73.4 }$ \footnotesize$\pm$ 1.4& $\mathbf{ 69.8 }$ \footnotesize$\pm$ 2.0 & $\mathbf{74.5}$\\

\midrule
{\textbf{VLCS 10-shot}} &                 C &           L &   V &        S & Average\\

ERM       &  99.5  \footnotesize$\pm$ 0.2&  52.6  \footnotesize$\pm$ 5.0&  74.8  \footnotesize$\pm$ 3.8&  69.1  \footnotesize$\pm$ 2.4 & 74.0\\ 
Ours       & $\mathbf{ 99.1 }$ \footnotesize$\pm$ 0.2& $\mathbf{ 65.0 }$ \footnotesize$\pm$ 6.2& $\mathbf{ 74.4 }$ \footnotesize$\pm$ 1.9& $\mathbf{ 70.8 }$ \footnotesize$\pm$ 2.3 & $\mathbf{77.3}$\\
\bottomrule
\end{tabular}

\label{tab:da_vlcs}

\end{table}

\begin{table}[h!]
\caption{results few-shot transfer learning on OfficeHome}
\centering
\begin{tabular}{cccccc}
\toprule
\multicolumn{2}{l}{\textbf{Dataset/Algorithm}} & \multicolumn{3}{c}{\textbf{OOD accuracy (by domain)}} \\
\midrule
{\textbf{OfficeHome 1-shot}} &         C &           A &   P &        R & Average\\

ERM       &  40.2  \footnotesize$\pm$ 2.4&  52.7  \footnotesize$\pm$ 2.6&  68.1  \footnotesize$\pm$ 1.7&  64.6  \footnotesize$\pm$ 1.8 & 56.4 \\

Ours       & $\mathbf{ 41.4 }$ \footnotesize$\pm$ 1.7& $\mathbf{ 54.5 }$ \footnotesize$\pm$ 2.0& $\mathbf{ 68.5 }$ \footnotesize$\pm$ 2.7& $\mathbf{ 69.0 }$ \footnotesize$\pm$ 1.5 & $\mathbf{58.4}$\\
\midrule

{\textbf{OfficeHome 5-shot}} &         C &           A &   P &        R & Average\\

ERM       &  63.2  \footnotesize$\pm$ 0.4&  73.3  \footnotesize$\pm$ 0.8&  84.1  \footnotesize$\pm$ 0.4&  82.0  \footnotesize$\pm$ 0.8 &  75.7\\

Ours       
& $\mathbf{ 66.2 }$ \footnotesize$\pm$ 1.2& $\mathbf{ 75.1 }$ \footnotesize$\pm$ 1.0& $\mathbf{ 83.6 }$ \footnotesize$\pm$ 0.5& $\mathbf{ 83.1 }$ \footnotesize$\pm$ 0.8 & $\mathbf{77.0}$\\
\bottomrule

{\textbf{OfficeHome 10-shot}} &         C &           A &   P &        R & Average\\

ERM       &  71.1  \footnotesize$\pm$ 0.4&  80.5  \footnotesize$\pm$ 0.5&  87.5  \footnotesize$\pm$ 0.3&  84.9  \footnotesize$\pm$ 0.5 & 81.0\\
Ours       & $\mathbf{ 72.2 }$ \footnotesize$\pm$ 1.2& $\mathbf{ 81.8 }$ \footnotesize$\pm$ 0.5& $\mathbf{ 87.5 }$ \footnotesize$\pm$ 0.2& $\mathbf{ 86.3 }$ \footnotesize$\pm$ 0.4 & $\mathbf{82.0}$ \\
\hline

\end{tabular}

\label{tab:da_offhome}
\end{table}

\begin{table}[h!]
\caption{results few-shot transfer learning Waterbirds}
\centering

\begin{tabular}{cccccc}
\toprule
\multicolumn{2}{l}{\textbf{Dataset/Algorithm}} & \multicolumn{3}{c}{\textbf{OOD accuracy (by domain)}} \\
\midrule
{\textbf{Waterbirds 1-shot}}&         LL &           LW &   WL &        WW & Average\\

ERM       &  99.1  \footnotesize$\pm$ 1.1&  43.8  \footnotesize$\pm$ 16.5&  79.5  \footnotesize$\pm$ 10.2&  86.7  \footnotesize$\pm$ 8.2 & 79.8\\
Ours      & $\mathbf{ 95.2 }$ \footnotesize$\pm$ 8.1& $\mathbf{ 81.9 }$ \footnotesize$\pm$ 9.5& $\mathbf{ 80.7 }$ \footnotesize$\pm$ 5.5& $\mathbf{ 95.9 }$ \footnotesize$\pm$ 1.2 & $\mathbf{88.4}$\\
\midrule

{\textbf{Waterbirds 5-shot}} &         LL &           LW &   WL &        WW & Average\\

ERM       &  96.3  \footnotesize$\pm$ 5.0&  58.7  \footnotesize$\pm$ 17.2&  80.1  \footnotesize$\pm$ 12.6&  84.1  \footnotesize$\pm$ 12.7 & 79.8\\
Ours       & $\mathbf{ 98.8 }$ \footnotesize$\pm$ 1.8& $\mathbf{ 75.4 }$ \footnotesize$\pm$ 9.0& $\mathbf{ 81.6 }$ \footnotesize$\pm$ 14.0& $\mathbf{ 94.8 }$ \footnotesize$\pm$ 1.8 & $\mathbf{87.6}$ \\
\midrule

{\textbf{Waterbirds 10-shot}} &         LL &           LW &   WL &        WW & Average\\

ERM       &  94.2  \footnotesize$\pm$ 4.2&  73.0  \footnotesize$\pm$ 11.6&  80.4  \footnotesize$\pm$ 6.3&  89.3  \footnotesize$\pm$ 3.3 & 84.2\\
Ours         & $\mathbf{ 98.2 }$ \footnotesize$\pm$ 0.9& $\mathbf{ 82.6 }$ \footnotesize$\pm$ 5.9& $\mathbf{ 80.7 }$ \footnotesize$\pm$ 6.3& $\mathbf{ 95.5 }$ \footnotesize$\pm$ 1.4 & $\mathbf{89.2}$ \\
\bottomrule

\end{tabular}

\label{tab:da_waterbirds}
\end{table}
\subsection{Feature sharing on \texttt{PACS}}

\begin{figure}[h!]
    \centering    \includegraphics[width=0.8\linewidth]{pictures/oty0iy1msketch0.01nex_150_mean_before__shared.png}  \\
    \includegraphics[width=0.8\linewidth]{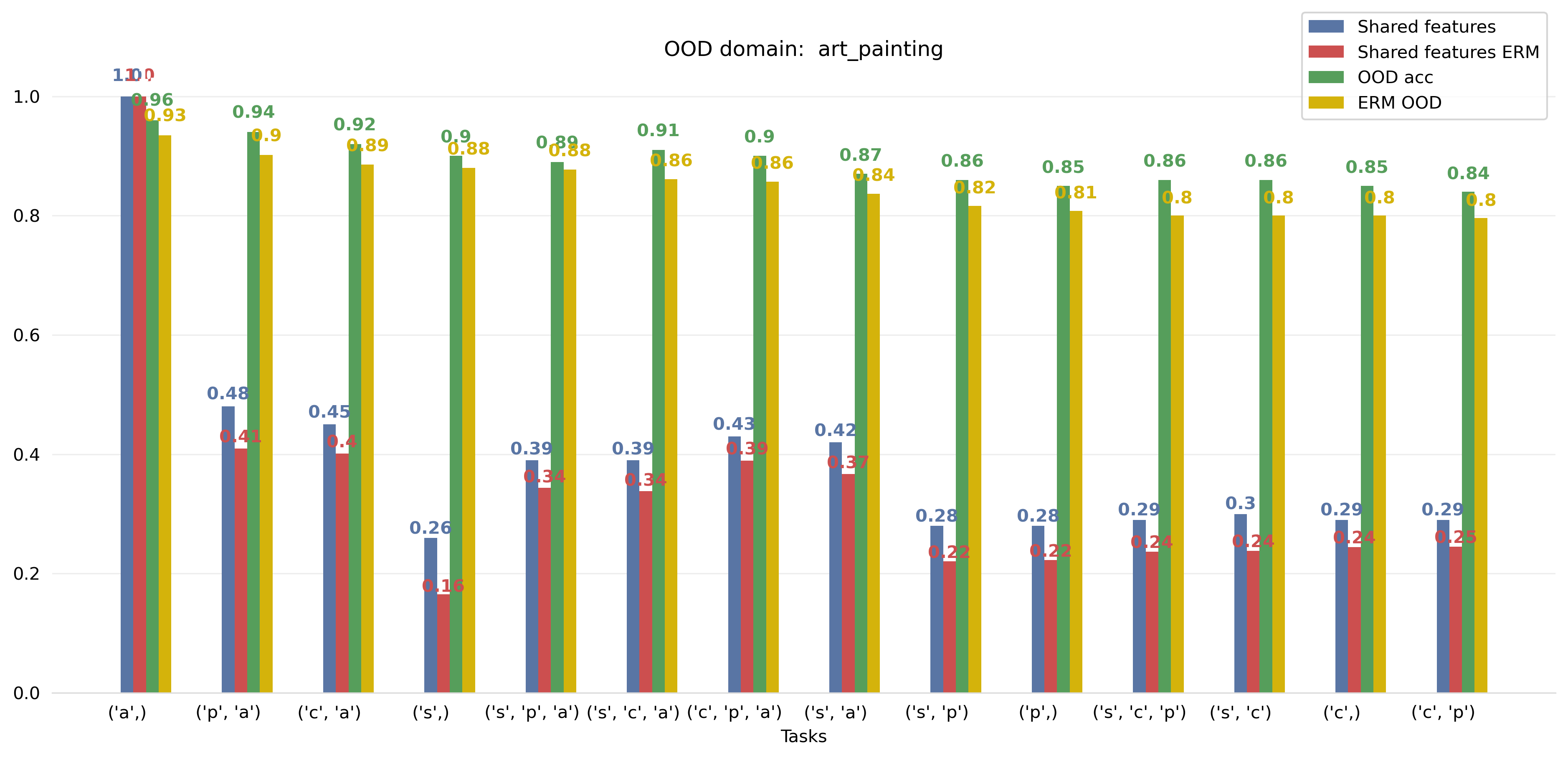}  \\
    \includegraphics[width=0.8\linewidth]{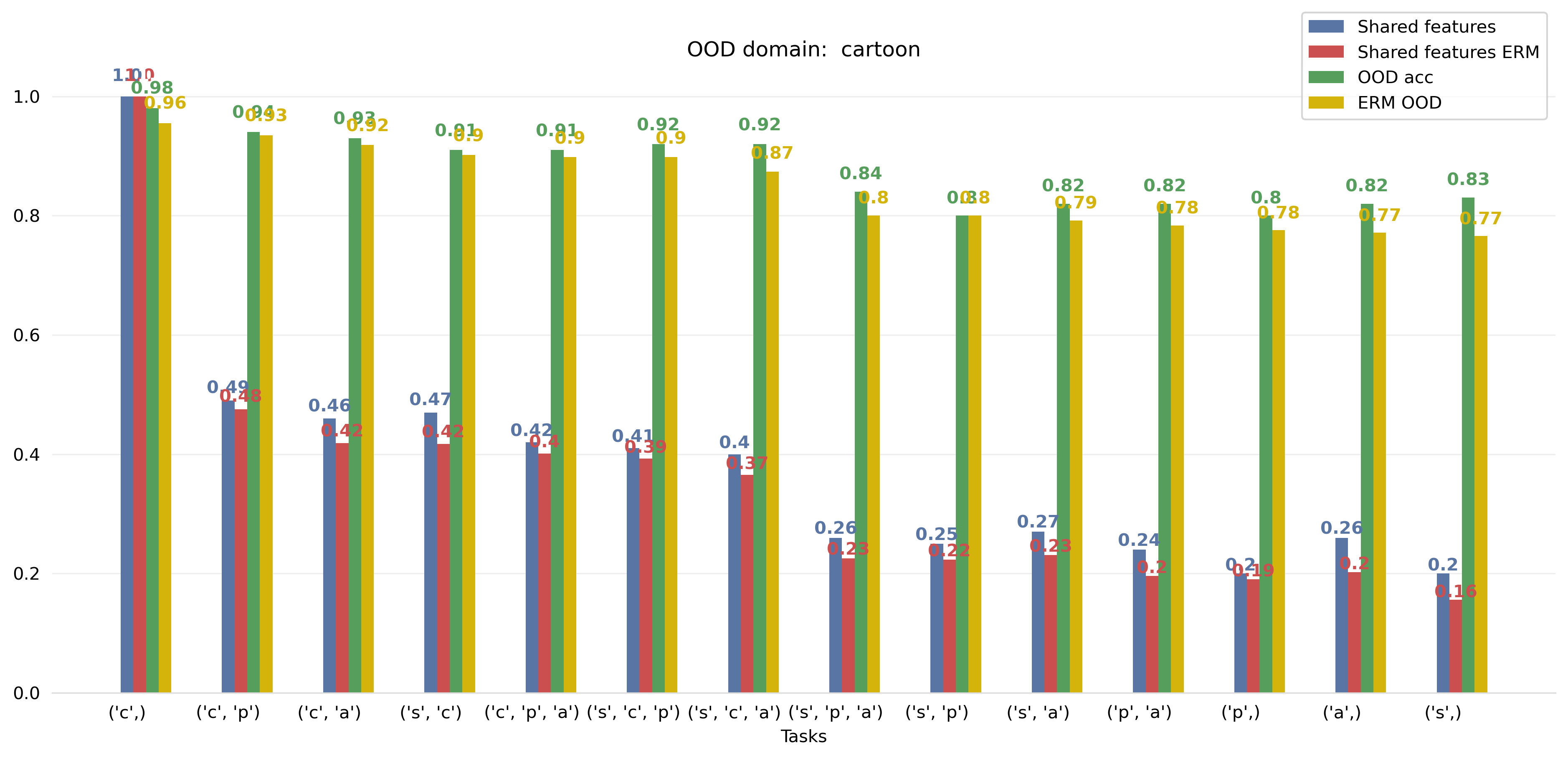}  \\
    \includegraphics[width=0.8\linewidth]{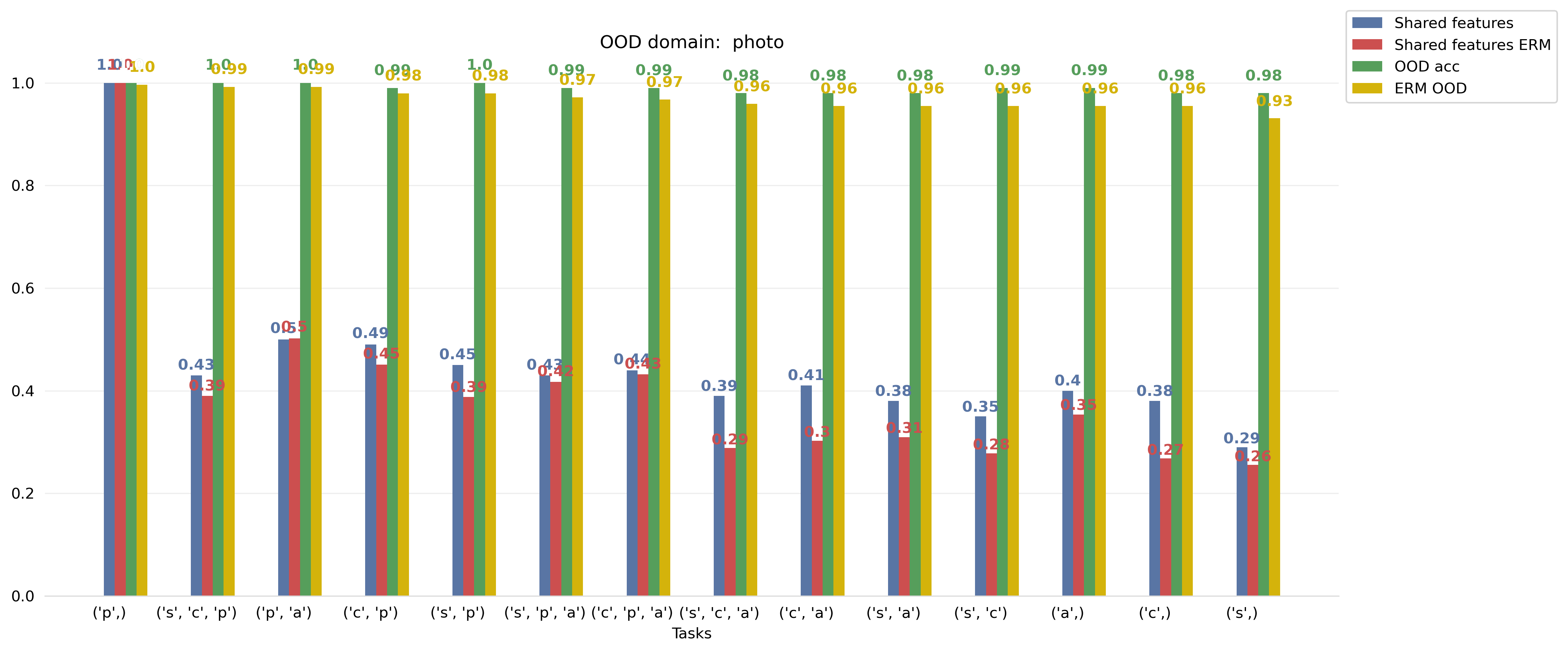} 
    \caption{Additional results for all domains in PACS, separated by domain. The overall message of Figure~\ref{fig:fract_shared_sketch} appear consistent across all domains.}
    \label{fig:pacs_all_feat_sharing}
\end{figure}

See Figure \ref{fig:pacs_all_feat_sharing} for additional results on all domains in \texttt{PACS}.

\subsection{Task similarity} \label{sec:app_task_sim} \label{sec:appendix_task_similarity_metric}
We show that our method enables direct extraction of a task representation and a metric for task similarity from our model and its feature space. We propose to use the coefficients of the fitted linear heads $f_{\phi_t^*}$ on a given task as a \emph{representation for that task}. Specifically we transform the optimal coefficients $\phi^*$ in a $M$-dimensional vector space (here $M$ is the number of features) by simply computing $\sum_c|\phi_{t,m,c}^*|$, and discretize them by a threshold $\epsilon$. The resulting binary vectors, together with a distance metric (we choose the Hamming distance), form a discrete metric space of tasks.
We preliminary verify how the proposed representation and metric behave on \texttt{MiniImagenet} \cite{vinyals2016matching} below.

\begin{figure}[h!]
\centering
\resizebox{\linewidth}{!}{
\begin{tabular}{lll}
\includegraphics[width=0.33\linewidth]{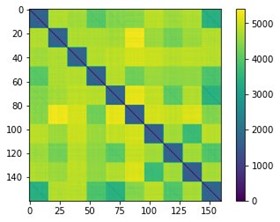}
& \includegraphics[width=0.32\linewidth]{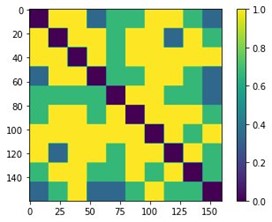}
& \includegraphics[width=0.35\linewidth]{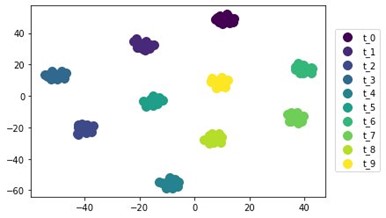}
\end{tabular}}
\caption{\emph{Task Similarity.} We visualize the tSNE of the discrete task representation and observe that the clusters in this space corresponds to group identities.}
\label{fig:task_similarity}
\end{figure}

We sample 160 tasks from 10 groups from , where each group has the same class support, i.e. $t_1,t_2 \in G_i\mapsto Supp(t_1)==Supp(t_2)  \forall i.$ We then fit the linear heads independently on each task (i.e. not using the feature sharing regularizer). Then we compute the discrete task representation and project the resulting vector space in a two dimensional vector space using tSNE \cite{wattenberg2016use}. The  clusters obtained in this space correspond exactly to the group identities (visualized in color in Figure \ref{fig:task_similarity}).

\subsection{Comparison with metalearning baselines} \label{sec:app_meta_learning}
In Table \ref{tab:meta_learning}, we further compare our method on meta learning benchmarks, namely \texttt{Mini Imagenet} \cite{vinyals2016matching} and {CIFAR-FS} \cite{bertinetto2018meta} with different approaches in the literature based on meta learning \cite{snell2017prototypical,oreshkin2018tadam,dhillon2019baseline,lachapelle2022synergies}.

In Figure \ref{fig:comparison_protonet} we compare the predicting performance of our method and capacity to leverage shared knowledge between task, comparing with backbone trained with protopical network approach. We sample a set of task with different overlap, where the overlap between two task $t_1,t_2$ is defined as $sim(t_1,t_2)=\frac{ Supp(t_1)\cap Supp(t_2)}{Supp(t_1)\cup Supp(t_2}$
indicating with $Supp(t_i)$ the support over classes in task $t_i$. We show that other than reaching a much higher accuracy the features of our model are able to be clustered at test time enabling to reach better performance on unseen task. As a matter of fact we can use the feature sharing regularizer at test time showing that there is  a increasing trend in the performance, while the prototypical networks features just decreases being unable to share information across tasks at test time.

\begin{table}[h!]
\centering
\caption{Meta learning baselines, including concurrent work~\cite{lachapelle2022synergies} which we significantly outperform.}
\resizebox{\linewidth}{!}{
\begin{tabular}{cccccc}
\toprule
                 & Architecture & Cifar-FS (1 shot) & Cifar-FS( 5 shot) & MiniImagenet(1 shot) & MiniImagenet (5 shot) \\
\midrule
MAML             & Conv32(x4)   &      -             &        -           & 48.7±1.84            & 63.11±0.66            \\
Prototypical Net & Conv64(x4)   &      -             &        -            & 49.42±0.78           & 68.20±0.66            \\
TADAM            & ResNet12     &      -             &        -           & 58.5 ±0.56           & 76.7 ±0.3             \\
MetaOptNet       & ResNet12     & 72.0 ± 0.7        & 84.2 ± 0.5        & $\mathbf{62.64}$±0.61           & $\mathbf{78.63}$±0.46            \\
MetaBaseline     & WRN 28-10    & $\mathbf{76.58}$±0.68        & 85.79±0.5         & 59.62 ±0.66          & 78.17 ±0.49           \\
\textit{Lachapelle et al}\cite{lachapelle2022synergies} & ResNet12     & -         & -        &  54.22 ± 0.6             & 70.01 ± 0.51          \\
Ours*            & ResNet12     & 75.1 ±0.4         & $\mathbf{86.9}$ ±0.19        & 60.1 ± 2             & 76.6 ± 0.1          \\
\bottomrule

\end{tabular}}
\label{tab:meta_learning}
\end{table}

\subsection{Sharing features at test time} \label{sec:app_feature_sharing_test_time}

Features can be enforced to be shared also at test time, simply by setting $\beta>0$ to fit the linear head on top of the learned feature space.
We observe the benefits of utilizing the feature sharing penalty at test time on the \texttt{Camelyon17} dataset in the fourth row of Table $\ref{tab:camelyon2}$.

 As highlighted in the main paper, retaining features which are shared across the training domains and cutting the ones that are domain-specific enable to perform better at test time, at the expenses of lower performance near the training distribution.

\begin{figure}[h]
\centering
\resizebox{\linewidth}{!}{
\begin{tabular}{lll}
\includegraphics[width=0.33\linewidth]{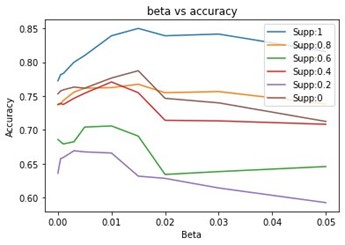}
& \includegraphics[width=0.33\linewidth]{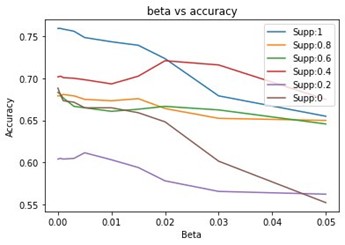}
& \includegraphics[width=0.33\linewidth]{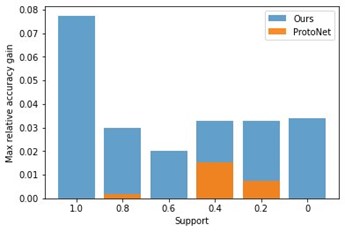}
\end{tabular}}
\caption{Enforcing feature sharing at test time. Our approach (on the left) is able to benefit from the feature sharing constraint at test time, while using the prototypical network backbone performance monotonically decrease (center). On the right we show the maximal performance gain for each group of tasks for the two approaches.}
\label{fig:comparison_protonet}
\end{figure}

 We analyzed in more depth this phenomenon in Figure \ref{fig:comparison_protonet}. For this experiment we trained our model and a Prototypical network \cite{snell2017prototypical} one on the \texttt{MiniImagenet} dataset. Then we sampled 5 groups of tasks according to an average overlap measure between tasks. Between two task $t_1,t_2$ the overlap is defined as $sim(t_1,t_2)=\frac{ Supp(t_1)\cap Supp(t_2)}{Supp(t_1)\cup Supp(t_2}$. each group is made of $10$ task.
 We then plot the performance at test time increasing the regularization parameter $\beta$, weighting the feature sharing.
 The outcome of the experiment is twofold: (i) we observe an increase in performance at test time, especially when tasks shows maximal overlap (i.e. they share more features) (ii) this is not the case with the pretrained backbone of \cite{snell2017prototypical} which shows almost monotonical decrease in the performance, i.e. enforcing the minimality property during training enables to use it as well at test time.

 Further analysis on different datasets, and also on tuning strategies on the regularization parameter are promising directions for future work, to better understand when and how enforcing feature sharing is beneficial at test time. 

\begin{table}[ht!]
    \caption{Camelyon17 quantitative results: we report accuracy both on ID and OOD splits. We show (last row) that feature sharing  at test time, leads to more robust features on OOD test data.}
    \centering
    \begin{tabular}{cccc}
    \toprule 
         & Validation(ID) & Validation (OOD) & Test (OOD) \\
    \midrule 
     ERM &  93.2          & 84               &  70.3      \\
     CORAL &  95.4        & 86.2             &  59.5     \\
     IRM &  91.6          & 86.2             &  64.2     \\
     Ours &  $\mathbf{93.2}$\footnotesize±0.3         & $\mathbf{89.9}$\footnotesize±0.6         &  74.1\footnotesize±0.2     \\
     Ours($\beta>0$ test) &  ${90.4}$\footnotesize±0.2    & 84.01\footnotesize±0.9        &  $\mathbf{85.5}$\footnotesize±0.6     \\
    \bottomrule 
    \end{tabular}
    \label{tab:camelyon2}
\end{table}

\end{document}